%% file: main.tex
\documentclass[runningheads]{llncs}

\usepackage[mobile]{eccv}

\usepackage{eccvabbrv}

\usepackage{graphicx}
\usepackage{booktabs}

\usepackage[accsupp]{axessibility}  

\usepackage{hyperref}

\usepackage{orcidlink}

\usepackage{framed}
\usepackage{color}
\usepackage{marvosym}
\usepackage[most]{tcolorbox}
\usepackage{xcolor}
\usepackage{colortbl}
\usepackage{mdframed}
\usepackage{changepage}
\usepackage{multirow}
\usepackage{multicol}
\usepackage{makecell}
\usepackage{amssymb}
\usepackage{pifont}
\newcommand{\cmark}{\ding{51}}%

\newcommand{\xhdr}[1]{{\noindent\bfseries #1}.}
\newcommand{\done}{\cellcolor{bubbles}}
\newcommand{\yr}{\textcolor{magenta}}

\definecolor{mygray}{gray}{.95}

\newenvironment{formal}{%
  \MakeFramed{\advance\hsize-\width\FrameRestore}%
  \noindent\hspace{-4.55pt}%
  \begin{adjustwidth}{}{}%
  \vspace{2pt}\vspace{-1pt}%
}
{%
  \vspace{-1pt}\end{adjustwidth}\endMakeFramed%
}

\definecolor{formalshade}{RGB}{213,230,242}
\definecolor{myblue}{RGB}{0, 77, 128}
\definecolor{bubbles}{rgb}{0.91, 1.0, 1.0}

\begin{document}


\title{\includegraphics[scale=0.025]{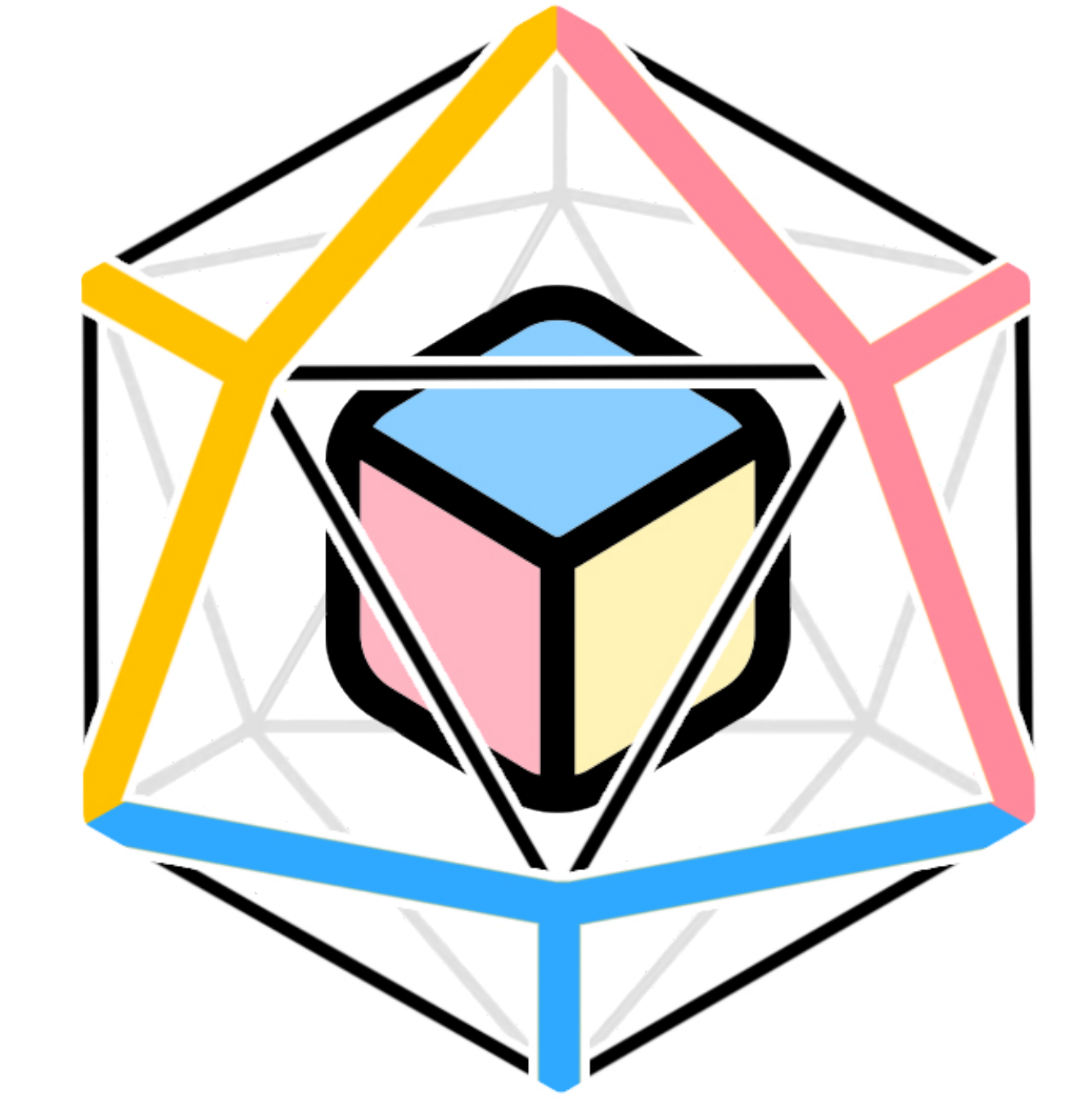} T$^3$Bench: Benchmarking Current Progress in Text-to-3D Generation} 

\author{
Yuze He\inst{1}$^*$ \and
Yushi Bai\inst{1}$^*$ \and
Matthieu Lin\inst{1} \and
Wang Zhao\inst{1} \and
Yubin Hu\inst{1} \and \\
Jenny Sheng\inst{1} \and
Ran Yi\inst{2} \and
Juanzi Li\inst{1} \and
Yong-Jin Liu\inst{1}\textsuperscript{\Letter}
}

\authorrunning{He Y., Bai Y. et al.}

\institute{Tsinghua University \and
Shanghai Jiao Tong University \\
{\tt\small \{hyz22,bys22\}@mails.tsinghua.edu.cn, liuyongjin@tsinghua.edu.cn}
  \phantom{\thanks{Equal contribution}} }

\newcommand{\ourmethod}{T$^3$Bench}

\maketitle

\input{sec/0_abstract}    
\input{sec/1_intro}

\input{sec/2_related}
\input{sec/3_method}
\input{sec/4_experiments}
\input{sec/5_conclusion}

%
%
\bibliographystyle{splncs04}
\bibliography{main}

\input{sec/X_suppl}

\end{document}

%% file: sec/0_abstract.tex
\begin{abstract}
Recent methods in text-to-3D leverage powerful pretrained diffusion models to optimize NeRF. Notably, these methods are able to produce high-quality 3D scenes without training on 3D data. Due to the open-ended nature of the task, most studies evaluate their results with subjective case studies and user experiments, thereby presenting a challenge in quantitatively addressing the question: How has current progress in Text-to-3D gone so far? In this paper, we introduce T$^3$Bench, the first comprehensive text-to-3D benchmark containing diverse text prompts of three increasing complexity levels that are specially designed for 3D generation. 
To assess both the subjective quality and the text alignment, we propose two automatic metrics based on multi-view images produced by the 3D contents. The quality metric combines multi-view text-image scores and regional convolution to detect quality and view inconsistency. The alignment metric uses multi-view captioning and GPT-4 evaluation to measure text-3D consistency. Both metrics closely correlate with different dimensions of human judgments, providing a paradigm for efficiently evaluating text-to-3D models. The benchmarking results, shown in Fig. 1, reveal performance differences among an extensive 10 prevalent text-to-3D methods. 
Our analysis further highlights the common struggles for current methods on generating surroundings and multi-object scenes, as well as the bottleneck of leveraging 2D guidance for 3D generation.
Our project page is available at: \url{https://t3bench.com}.
\keywords{Text-to-3D \and Evaluation \and Generative Models}
\end{abstract}

%% file: sec/1_intro.tex
\section{Introduction}
\label{sec:intro}

\vspace{-5mm}
\begin{formal}
\textit{It is a narrow mind which cannot look at a subject from various points of view. \hfill --- George Eliot}
\end{formal}
\vspace{-2mm}

Equipping machines with the ability to automatically generate 3D objects and scenes from text descriptions has long been an ambitious and ongoing pursuit.
Recent methods, such as diffusion model~\cite{ho2020denoising,rombach2022high} and NeRF~\cite{mildenhall2021nerf, zhang2020nerf++, he2023mmpi}, have significantly improved the effectiveness of text-to-3D methods, empowering potential applications ranging from arts realization to industrial design.

\begin{figure}[t]
\centering
\includegraphics[width=1.0\linewidth]{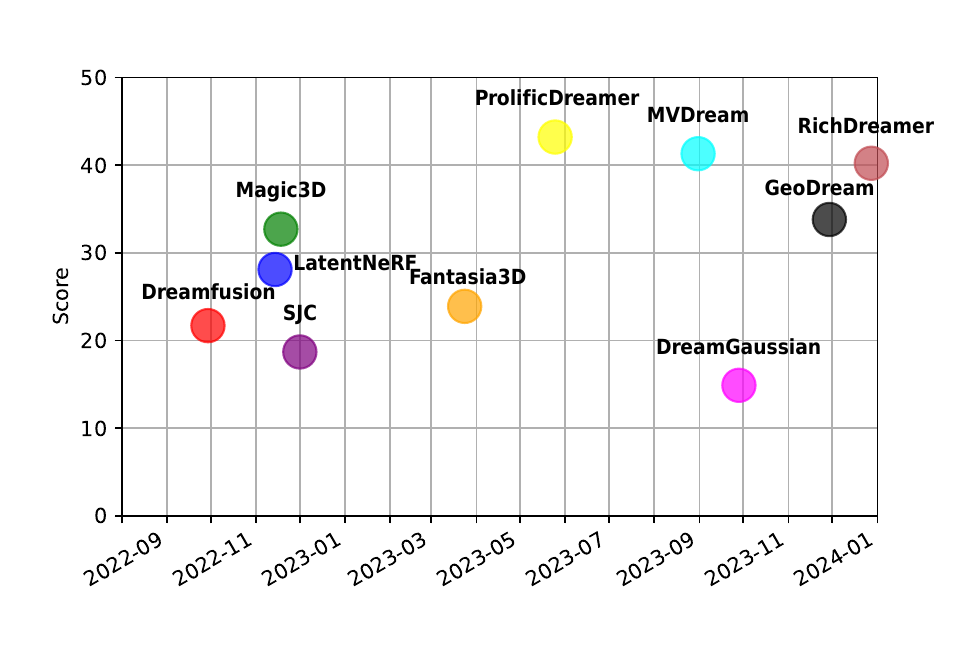}
\vspace{-30pt}
\caption{The average scores of 10 prevalent text-to-3D methods on \ourmethod, computed by the mean of quality \& alignment metrics.}
\label{fig:resolution}
\vspace{-20pt}
\end{figure}

However, there lacks a systematic approach to benchmarking current progress on text-to-3D methods, which is most prominently reflected in two aspects:
\textcolor{myblue}{(a)} A lack of a standard set of diverse, challenging test textual inputs.
\textcolor{myblue}{(b)} An absence of a set of automatic and comprehensive evaluation metrics to quantitatively measure the quality of the generated 3D scenes.
Specifically, previous works~\cite{seo2023let, lin2023magic3d, seo2023ditto} mostly adopt simple object or scene prompts for evaluation, and largely rely on subjective user experiments.
Several works~\cite{tsalicoglou2023textmesh, poole2023dreamfusion, mohammad2022clip, xu2023dream3d} assess 3D generation quality by rendering the generated 3D model into a single 2D image and measuring its alignment with the text prompt through CLIP cosine distance or CLIP R-precision.
Nevertheless, they only consider \textbf{one view} of the 3D scene, failing to assess the overall 3D quality.

\begin{figure*}[h]
\centering
\includegraphics[width=0.93\linewidth]{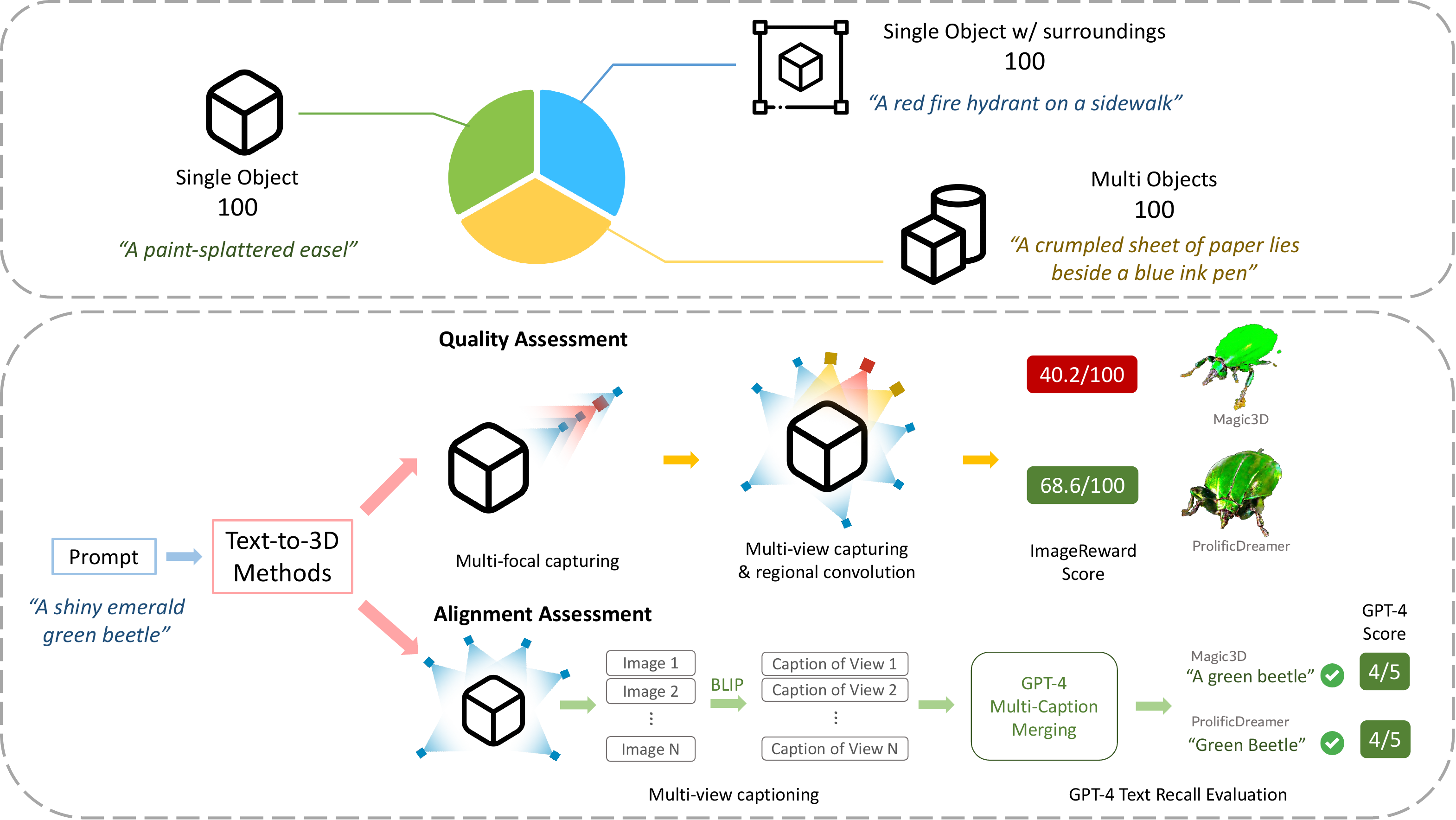}
\caption{The overview of our \ourmethod \ benchmark pipeline. }
\label{fig:detail}
\end{figure*}

To facilitate further research in this direction, we introduce \ourmethod, the first comprehensive text-to-3D benchmark.
For a careful and thorough assessment, we build the benchmark to accurately reflect the primary challenges of current text-to-3D approaches. This includes their scalability and robustness in generating a variety of 3D scenes, the quality and view consistency of these generated scenes, and the correctness or alignment of these 3D scenes with their respective texts.
Specifically, we devise three prompt suites incorporating diverse 3D scenes and with increasing complexity, including {\it Single object}, {\it Single object with surroundings}, and {\it Multiple objects}.
We also propose two automatic evaluation metrics that both take \textbf{multi-view information} into consideration,
focusing on assessing the subjective quality of the generated 3D scenes and its alignment with the textual prompt respectively.
To calculate these two metrics, we first employ multi-focal and multi-view capturing to obtain a set of 2D images from the generated 3D scenes.
The {\it quality} metric individually scores these multi-view images with text-image scoring models (CLIP~\cite{radford2021learning}, ImageReward~\cite{xu2023imagereward}), and then combines them into one overall quality measurement using {\it regional convolution}, which also effectively detects the infamous Janus problem (view inconsistency) in prevalent text-to-3D models~\cite{poole2023dreamfusion,hong2023debiasing}.
On the other hand, the {\it alignment} metric utilizes multi-view captioning and GPT-4 evaluation to measure how closely the 3D information aligns with the textual information in the input text prompt.
Our user experiments show that both metrics correlate closely with human scorings in 1-5 scale (with a Spearman correlation higher than 0.75), demonstrating them as efficient and accurate measurements.

As the \textbf{first} attempt to benchmark current text-to-3D methods, \ourmethod\ yields fruitful results.
Our benchmark reveals the strengths and weaknesses across 10 prevalent text-to-3D methods, as well as their common insufficiency when faced with more complicated 3D scenes, such as those involving multiple objects.
We also analyze the correlation between the performance of text-to-3D methods and the quality of the 2D guidance generated by diffusion models, showing that the primary hurdle for text-to-3D mainly lies in the transition from 2D to a consistent 3D scene.

%% file: sec/2_related.tex
\section{Related Works}

\xhdr{Text-to-3D}
Predominant works in text-to-3D~\cite{poole2022dreamfusion, lin2023magic3d, qian2023magic123, chen2023fantasia3d, wang2023prolificdreamer, metzer2022latentnerf} circumvent the need for 3D training data by using large pretrained text-to-image diffusion models~\cite{stablediffusion, imagen}.
However, these approaches suffer from inconsistency between views.
Notably, the proposed score distillation loss~\cite{poole2022dreamfusion} does not take into account the consistency between views as the diffusion model mimics a stochastic process~\cite{ho2020denoising}. On the one hand, ProlificDreamer~\cite{wang2023prolificdreamer} proposes a variational formulation of the score distillation loss to consider the stochasticity in the diffusion process. On the other hand, some researchers propose to fine-tune the diffusion model to improve its consistency across views~\cite{efficientdreamer}. 
However, current metrics do not adequately consider the 3D nature of the generated results, which makes it difficult to compare the effectiveness of different methods. Prior work has relied either on labor-intensive user studies~\cite{wang2023prolificdreamer} or CLIP R-precision~\cite{radford2021learning}, which does not consider 3D consistency.
While early attempts have been made to measure 3D consistency~\cite{hong2023debiasing}, these efforts only capture one aspect of the problem and overlook crucial metrics such as quality and prompt alignment.

\xhdr{Text-to-image Generation and Evaluation} With the development of diffusion models~\cite{ho2020denoising}, text-based image generation has experienced significant progress in recent years~\cite{stablediffusion, imagen}. These models excel at complex tasks like editing and composition~\cite{hertz2022prompt, brooks2023instructpix2pix}. However, comparing their capabilities in text-based generation is challenging due to the open-ended nature of the task~\cite{bakr2023hrs}. Prior work in text-to-image generation introduced DrawBench~\cite{imagen}, a comprehensive set of prompts aiming to evaluate various aspects, including color understanding, object recognition, and spatial relations. Other approaches leverage CLIP~\cite{radford2021learning} and BLIP~\cite{li2022blip} to measure the similarity between text and generated images by using these models as scorers to gauge prompt alignment. In a similar vein, the Aesthetic score~\cite{schuhmann2022laion} employs the CLIP model to predict image aesthetics. While these methods assess alignment and quality to some extent, they fall short in considering multiple properties like toxicity, quality, and alignment. To encompass these diverse properties into a single model,  ImageReward~\cite{xu2023imagereward} proposes training a reward model via reinforcement learning from human feedback. Results show that this reward model better aligns with human preferences.
Although evaluation for 3D generation can draw on text-to-image evaluation methods, it is important to note the major difference between the two: 3D contains semantic information from multiple viewpoints rather than a single view.

%% file: sec/3_method.tex
\section{Method}
\label{sec:method}

This section presents
the methodology in constructing \ourmethod, including the design and generation of text prompts, the unification of 3D representations, and the introduction of two novel evaluation metrics --- the quality assessment and the alignment assessment.

\subsection{Diverse Prompts with Increasing Complexity}
\label{sec:prompt}

While there are some widely used text-to-image prompt sets, such as DrawBench~\cite{saharia2022photorealistic} and DALL-EVAL~\cite{cho2023dalleval}, many of the prompts in these benchmarks pose substantial challenges for existing text-to-3D methods and lack an adequate degree of distinction.
Certain prompts, for instance, are excessively lengthy, while others incorporate complex aspects such as counting, leading to poor 3D scenes generated by all current text-to-3D methods.
Therefore, a new set of prompts needs to be specifically crafted for evaluating prevalent text-to-3D methods.

We observe that current text-to-3D approaches demonstrate relatively robust performance on prompts with a single object. However, their performance notably declines on text prompts that include environmental surroundings or multiple objects. Such deficiency is partly due to the utilization of 2D supervision, which cannot ensure consistency amongst different viewpoints. 
With these observations, we design three prompt sets with increasing complexity to perform a targeted evaluation of text-to-3D approaches, namely {\it Single object}, {\it Single object with surroundings}, and {\it Multiple objects}. The {\it Single object} set represents the simplest scenario to establish a baseline level of performance, and the other two prompt sets introduce increased levels of difficulty by incorporating additional information, {\it i.e.,} surroundings or multiple objects.

To generate these prompt sets, we first use GPT-4~\cite{openai2023gpt4} to generate a large pool of candidate prompts, and then manually filter out prompts that contain proper nouns or toponyms. Subsequently, we utilize ROUGE-L~\cite{lin2004rouge} to quantify prompt similarity and
gradually remove highly similar prompts until there remains a number of $N$ distinct prompts with significant diversity in each prompt set.

\subsection{Unified 3D Representation}

Different text-to-3D methods employ various 3D representations during generation, such as  NeRF~\cite{mildenhall2021nerf} and 3D mesh. 
From a testing perspective, a 3D mesh is more conducive than NeRF due to its explicit geometric structure, which facilitates localization and normalization. Moreover, the primary use of text-to-3D is to obtain editable 3D assets that can be applied in fields such as virtual reality and gaming. Considering the purpose and practical applications, 3D mesh is a more suitable unified representation for benchmarking text-to-3D methods.
We convert NeRF generated by text-to-3D methods
into a 3D mesh using either DMTet~\cite{shen2021deep} or Marching Cube~\cite{lorensen1998marching},
and choose the one that produces superior results. 
This makes subsequent evaluations more convenient while encourages the generation of 3D scenes with more compact and clear geometry.

\subsection{Evaluation Metrics on Quality and Alignment}
\label{sec:metric}

\subsubsection{Overview}
The evaluation of text-to-3D methods remains challenging due to the need to fully account for the quality, view consistency, and text alignment of the generated 3D scenes.

Our evaluation metrics primarily focus on two dimensions~\cite{lee2024holistic} that typically reflect the effectiveness of text-to-3D methods: (1) the subjective quality of the generated 3D scene, and (2) the degree of alignment between the generated 3D scene and the input text prompt. 
To assess quality, we devise a scoring mechanism that comprises multi-focal and multi-view capturing, and utilizes text-image scoring models to obtain an overall quality measurement of the generated 3D scene. 
As for the textual alignment, we develop a scoring metric based on multi-view captioning and GPT-4 evaluation. 

\subsubsection{Quality Assessment}

Since the spatial geometry information is crucial for the generated 3D scenes, evaluation from a single view is incapable of assessing the quality of the generated results.
We believe a comprehensive and reliable 3D quality assessment should take into account the following aspects:
(a) \emph{Viewpoint selection}: choosing an appropriate viewpoint can better reflect the quality of the 3D scene, particularly potential object occlusions;
(b) \emph{Area coverage}: it is essential to simultaneously examine the current viewpoint and adjacent areas.
By doing so, the assessment can take into account a more global geometry, thereby avoiding a collapse to a local optimal view that leads to failure in detecting 3D consistency issues like the Janus problem.

To meet these requirements,
we incorporate a delicate
capturing and scoring procedure to evaluate the quality of the 3D generation. The following steps outline our method:

\begin{figure}[t]
\centering
\includegraphics[width=0.7\linewidth]{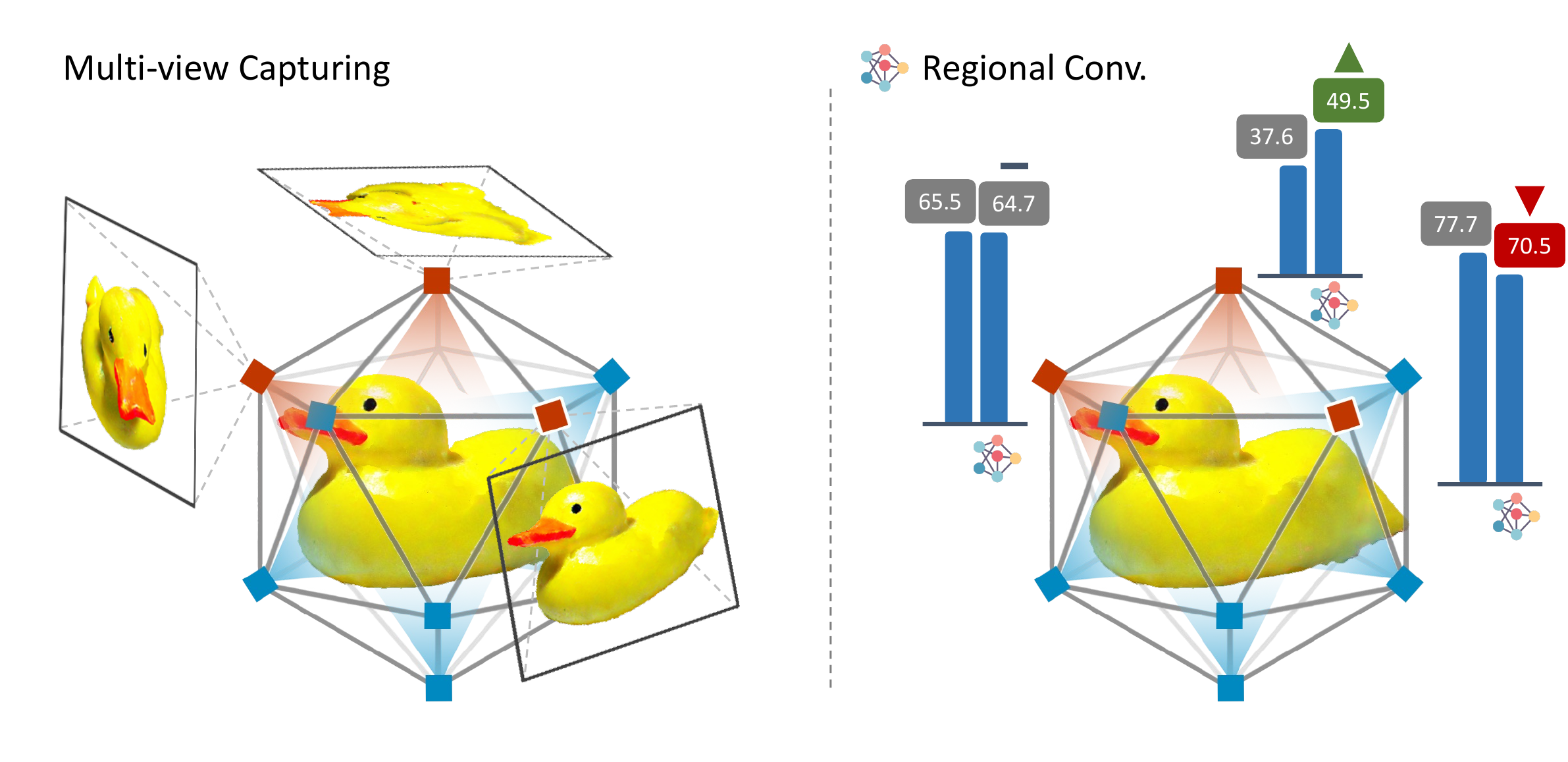}
\caption{Demonstration of scores at different viewpoints after multi-view capturing and regional convolution. Here, we use a level-0 icosahedron for a schematic illustration, please refer to Fig.~\ref{fig:ico} in the supplementary material for more details.}
\label{fig:conv}
\vspace{-20pt}
\end{figure}

\xhdr{Mesh Normalization} 
We convert the generated 3D scene into a mesh and scale it proportionally in the $x$, $y$, and $z$ directions, allowing the mesh to fit within a cube with a range of $[-1, 1]$ on all three dimensions. This helps to roughly determine the mesh's range for subsequent capturing.

\xhdr{Multi-Focal Capturing}
Capturing a 2D image using a fixed focal length from a single location may yield inaccurate evaluation results.
This is because the information in the captured image may be incomplete when the focal length is too long, and may occupy only a small portion in the frame when the focal length is too short.
To address this issue, we employ five different focal lengths to capture the mesh at each location and select the best focal length based on the highest text-image score.

\xhdr{Multi-View Capturing} 
To capture the 3D scene as completely as possible, we construct an icosahedron with a radius of $2.2$ around the origin and capture the 3D scene from all the vertices of the icosahedron (see an illustration of icosahedron in Fig.~\ref{fig:conv},\ref{fig:ico}). As text-image scoring models may be sensitive to rotation, we ensure that the plane formed by the up vector and look-at vector during capture contains the vertical axis. In practice, we use a level-2 icosahedron and capture from 161 locations.

\xhdr{Scoring and Regional Convolution} 
We employ text-image scoring models, e.g., CLIP~\cite{radford2021learning} and ImageReward~\cite{xu2023imagereward}, to score the 2D views from all 161 icosahedron vertices along with the textual prompt.
To capture a more global feature, we apply a pooling operator to the score at each location.
Standard averaging of scores across all locations may not be appropriate, as most views are not suitable for evaluation (e.g., top or bottom), and this approach may oversmooth the actual performance.
Meanwhile, taking the overall maximum scores may overlook the view inconsistency issue.
Therefore, we design a {\it regional convolution mechanism} to smooth out the score over each local region.
We treat the icosahedron as a graph composed of vertices and edges, and perform mean pooling on the graph with the following recursive formula:
\begin{equation}
s_{i}^{(t+1)} = \frac{1}{w|N(i)|+1} \left( s_{i}^{(t)} + w\sum_{j \in N(i)} s_{j}^{(t)} \right),
\end{equation}
\noindent
where $s_{i}^{(t)}$ is the score of point $i$ on the icosahedron at the $t$-th iteration, $N(i)$ is the set of neighboring points of $i$, and $|N(i)|$ is the number of neighbors of $i$. The superscript $(t+1)$ denotes the score after the $(t+1)$-th iteration.
We choose a total of $t=3$ iterations of mean pooling and convolution weight $w=1$ as we empirically find that they ensure a balance between adequate smoothing and over-smoothing
(please refer to Sec.~\ref{sec:conv} for more details).

After these steps, we select the highest score from all viewpoints as the final quality score for the 3D generation.

\subsubsection{Alignment Assessment}

In addition to the evaluation from the quality aspect, the alignment between 3D semantic information and text is another crucial aspect that should be considered. 
To measure the alignment between different modalities, we first perform 3D-to-text
captioning on the 3D scene and then compute the similarity between the caption and the textual prompt.

Directly utilizing image captioning methods such as BLIP~\cite{li2022blip} on a single view may fail to reflect the comprehensive information of a 3D object. 
To this end, we utilize a 3D-to-text captioning pipeline similar to Cap3D~\cite{luo2023scalable}. Initially, a level-0 icosahedron consisting of 12 vertices is established around the origin. This icosahedron captures the 3D scene on the 12 locations, each of which is captioned using BLIP. We then employ GPT-4~\cite{openai2023gpt4} to merge these captions (detailed in Sec.~\ref{ssec:caption_merging} of supplementary material), resulting in a 3D caption for the object.

Upon obtaining the 3D caption, we need to measure its alignment with the original prompt, particularly concerning the \emph{recall} of the original prompt within the caption.
Specifically, we observe that the text-to-3D methods might generate features not mentioned in the prompt (e.g., a red beak feature on a rubber duck), which may be reflected in the caption provided by BLIP.
Such additional features should not be considered misalignments, even though many similarity-based scoring methods (BLEU, BERTScore~\cite{zhang2019bertscore}) might assign them lower scores.
To assess the text recall, we adopt ROUGE-L~\cite{lin2004rouge}.
We also incorporate GPT-4 as text recall evaluators, drawing upon their demonstrated ability to mimic human experts in data annotation and evaluation~\cite{bai2023benchmarking}.
Here is the prompt we use:

\begin{tcolorbox}[size=title,opacityfill=0.1]
\noindent 
\textbf{Prompt:}
You are an assessment expert responsible for prompt-prediction pairs. Your task is to score the prediction according to the following requirements:

1. Evaluate the recall, or how well the prediction covers the information in the prompt. If the prediction contains information that does not appear in the prompt, it should not be considered as bad. 

2. If the prediction contains correct information about color or features in the prompt, you should also consider raising your score.

3. Assign a score between 1 and 5, with 5 being the highest. Do not provide a complete answer; give the score in the format: 3

Prompt: A photographer is capturing a beautiful butterfly with his camera

Prediction: A man photographing a butterfly near a tree and map, surrounded by plants
\\
\textbf{Answer: }
 \underline{4}
\end{tcolorbox}

%% file: sec/4_experiments.tex
\section{Experiments}

\subsection{Metric Evaluation}

In order to validate the reliability of our proposed metrics, we conduct a human-centered evaluation.
Expert evaluators are tasked with manually assigning scores to 3D scenes generated by 7 different methods (detailed in Sec.~\ref{ssec:benchmark_results}) on 30\% of all the prompts in \ourmethod. This results in a total of 1,260 scores. The human annotations span two dimensions: quality, which concerns the subjective quality of the generated results, and alignment, which focuses on the extent to which the generated content covers the original prompt. These evaluations are quantified using a 1-5 Likert scale. Subsequently, we measure the correlation between the proposed metrics and human annotations using Spearman's $\rho$, Kendall's $\tau$, and Pearson's $\rho$ correlation coefficients.

\begin{table}[h]
\centering
\vspace{-10pt}
\caption{Overview of evaluation metrics used in previous works.}
\label{tab:otherworks}
\begin{tabular}{lccc}
\toprule
& User Study & \quad CLIP R-Precision \quad & CLIP Similarity \\
\midrule
Dream Fields\cite{jain2022zero} & &\cmark & \\
DreamFusion\cite{poole2023dreamfusion} & & \cmark & \\
Magic3D\cite{lin2023magic3d} & \cmark & & \\
ProlificDreamer\cite{wang2023prolificdreamer} & \cmark & & \\
MVDream\cite{shi2023mvdream} &  \cmark & & \\
GaussianDreamer\cite{yi2023gaussiandreamer} & & & \cmark \\
Instant3D\cite{li2023instant3d} & & \cmark & \\
\bottomrule
\end{tabular}
\label{tb:metric}
\vspace{-10pt}
\end{table}

\newcommand*\mc[1]{\multicolumn{2}{c}{\textbf{#1}}}

\begin{table*}[h]
\centering  
\caption{The correlation of different combinations of evaluation methods with human annotations.
}
\resizebox{\linewidth}{!}{
\begin{tabular}{lcccccc}
\toprule[2pt]
& \cellcolor{mygray}CLIP & \cellcolor{mygray}CLIP & \mc{\hspace{2mm} Multi-view Quality} & \mc{\hspace{1mm} Multi-view Alignment} \\
& \cellcolor{mygray}Similarity & \cellcolor{mygray}R-Precision & \hspace{4mm} CLIP & ImageReward & \hspace{4mm} ROUGE-L & GPT-4 \\
\midrule

\emph{Quality} \\
Spearman ($\rho$) \yr{$\uparrow$} & \cellcolor{mygray}0.638 & \cellcolor{mygray}0.464 & \hspace{4mm} 0.749 & \textbf{0.784} & \hspace{4mm} 0.407 & 0.593 \\ 
Kendall ($\tau$) \yr{$\uparrow$} & \cellcolor{mygray}0.496 & \cellcolor{mygray}0.420 & \hspace{4mm} 0.597 & \textbf{0.636} & \hspace{4mm} 0.310 & 0.508 \\ 
Pearson ($\rho$) \yr{$\uparrow$} & \cellcolor{mygray}0.621 & \cellcolor{mygray}0.457 & \hspace{4mm} 0.727 & \textbf{0.780} & \hspace{4mm} 0.393 & 0.554 \\ 
\midrule
\emph{Alignment} \\
Spearman ($\rho$) \yr{$\uparrow$} & \cellcolor{mygray}0.638 & \cellcolor{mygray}0.479 & \hspace{4mm} 0.745 & 0.722 & \hspace{4mm} 0.567 & \textbf{0.780} \\ 
Kendall ($\tau$) \yr{$\uparrow$} & \cellcolor{mygray}0.495 & \cellcolor{mygray}0.433 & \hspace{4mm} 0.590 & 0.572 & \hspace{4mm} 0.442 & \textbf{0.701}  \\ 
Pearson ($\rho$) \yr{$\uparrow$} & \cellcolor{mygray}0.627 & \cellcolor{mygray}0.475 & \hspace{4mm} 0.730 & 0.713 & \hspace{4mm} 0.574 & \textbf{0.774} \\ 

\bottomrule[2pt]
\end{tabular}
}
\label{tb:corr}
\vspace{-20pt}
\end{table*}

We list the commonly used metrics for text-to-3D evaluation in Table~\ref{tb:metric}. The CLIP R-Precision and CLIP Similarity metrics used in previous works only consider one view, and we compare them with our proposed multi-view based metrics. Specifically, for our quality metric, we consider CLIP~\cite{radford2021learning} and ImageReward~\cite{xu2023imagereward} as single-view scoring methods; for the alignment metric, we explore the use of ROUGE-L and GPT-4 to measure text recall.
CLIP R-Precision uses CLIP to retrieve the correct caption among a set of distractors given a rendering of the content. Following prior works\cite{poole2022dreamfusion, jain2022zero}, we use the 153 prompts from the object-centric COCO\cite{lin2014microsoft} validation subset of Dream Fields\cite{jain2022zero} as the negative prompt set.

We report the results in Tab.~\ref{tb:corr}.
Drawing conclusion from the first four columns, we validate that our proposed multi-view based metrics are superior to single-view examination.
Moreover, compared to ROUGE-L, GPT-4 provides a more reliable assessment of alignment, as depicted by the last two columns.
These findings justify the design of our processing and scoring methods in Sec.~\ref{sec:metric}.
The inherent characteristics of retrieval-based metrics, which provide assessments that are comparative rather than absolute, also result in misalignment with human perceptual processes.
Overall, we observe that Multi-view capturing + ImageReward and 3D captioning + GPT-4 scoring align most closely with quality and alignment aspects as annotated by human experts, respectively.
We thus employ these combinations as the default quality and alignment metrics in our benchmark, throughout the rest of the paper.

\xhdr{Janus problem analysis}
The Janus problem, or multi-view inconsistency issue, arises when using Stable Diffusion for guidance, as it may not always generate accurate front, side, or back views for training. Consequently, this can lead to the regeneration of content described by the text prompt, and the most canonical view (e.g., the front piggy face) of an object appears in other views (e.g., the back view of a piggy).
In the following, we validate that \textbf{3D scenes with the Janus problem can be reflected in our multi-view metrics}.

\begin{figure*}[htb]
\vspace{-20pt}
\centering
\includegraphics[width=1.0\linewidth]{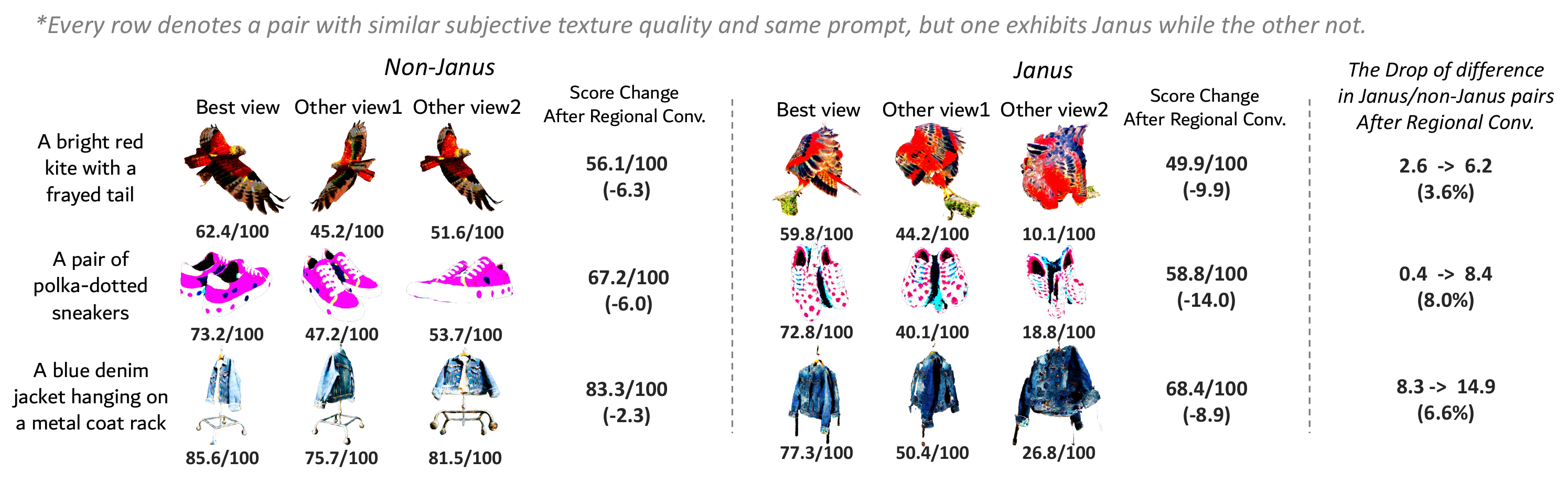}
\caption{
Underlying mechanism of how our multi-view quality metric reflects the Janus problem: Scores for illed-views are penalized, and regional convolution propagates this drop in local score to the global score.
}
\label{fig:janusplus}
\vspace{-40pt}
\end{figure*}

Intuitively, given a large number of viewpoints, for an object with the Janus problem, many views will show wrong results and lead to a decline in the quality score. After employing regional convolution to evaluate the quality of a more global region, our multi-view quality metric is able to faithfully reflect the Janus problem within the generated 3D scenes. This mechanism is illustrated in Fig.~\ref{fig:janusplus}.

\begin{figure}[htb]
\centering
\begin{minipage}[t]{0.48\textwidth}
    \centering
    \vspace{-10pt}
    \captionof{table}{Relative quality score drop from 3D scenes without Janus problem to scenes with Janus problem.}
    \vspace{10pt}
    \resizebox{\linewidth}{!}{
    \begin{tabular}{lcc}
    \hline
     & \makecell{Relative score drop \\w/o regional conv} & \makecell{Relative score drop\\w/  regional conv} \\
    \hline
    \makecell[c]{\textbf{Randomly} \\ \textbf{Selected Pairs}} & 2.93\% & \textbf{4.45\%} \\
    \makecell[c]{\textbf{Artificially} \\ \textbf{Modified Pairs}} & 4.05\% & \textbf{7.01\%} \\
    \hline
    \end{tabular}
    }
    \label{tab:januspair}
\end{minipage}
\quad
\begin{minipage}[t]{0.48\textwidth}
    \centering
    \caption{The quality score (normalized to 0-100) distribution of generated 3D scenes with and without Janus problem.}
    \includegraphics[width=1\linewidth]{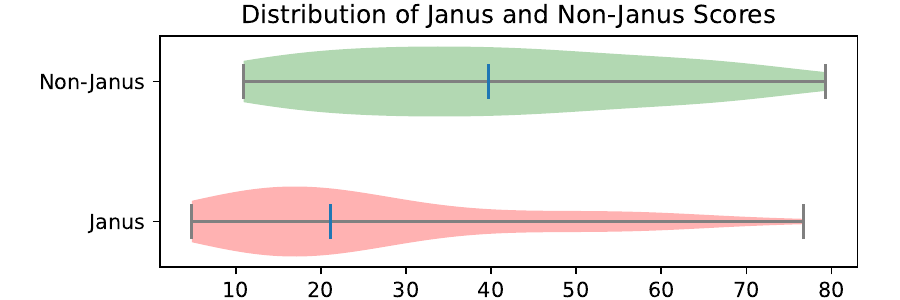}
    \label{fig:violin}
\end{minipage}
\vspace{-30pt}
\end{figure}

We perform two evaluations to investigate the discrepancy in scores for the Janus problem. 
The first is by randomly selecting 30 pairs of results generated by two different methods using the same prompts and with similar texture quality
(that is indistinguishable upon human examination),
meanwhile one with and the other without the Janus problem. 
Secondly, we take 15 generated scenes without the Janus problem and artificially synthesized scenes with the Janus problem by rotating counterparts and fusing them. 
For both cases, we compare the changes in the quality metric score before and after applying regional convolution. The results in Table \ref{tab:januspair} show a clear discrepancy in scores, especially after applying regional convolution.

We also present the quality score trend on randomly selected 70 non-collapsed meshes and categorize them based on the presence of the Janus Problem (shown in Fig.~\ref{fig:violin}). We observe that 3D scenes with the Janus Problem are generally scored lower.

\subsection{Benchmarking Results}
\label{ssec:benchmark_results}

\xhdr{Experimental Setup} Following the prompt generation scheme outlined in Sec.~\ref{sec:prompt} and taking into consideration both experimental breadth and test speed, we utilize GPT-4 to generate $N=100$ prompts for each of the three categories: {\it single object}, {\it single object with surroundings}, and {\it multiple objects}, resulting in a total of 300 prompts. 
We employ the implementation provided by ThreeStudio~\cite{threestudio2023} (or its extensions) to uniformly evaluate 10 prevalent text-to-3D methods on these prompts, including DreamFusion~\cite{poole2023dreamfusion}, Magic3D~\cite{lin2023magic3d}, LatentNeRF~\cite{metzer2022latentnerf}, Fantasia3D~\cite{chen2023fantasia3d}, SJC~\cite{wang2023score}, ProlificDreamer~\cite{wang2023prolificdreamer}, MVDream~\cite{shi2023mvdream},
DreamGaussian~\cite{tang2023dreamgaussian},
GeoDream~\cite{ma2023geodream},
and RichDreamer~\cite{qiu2023richdreamer}.
We normalize the original scores on quality and alignment assessment from the range $[-2.5, 2.5]$, $[1, 5]$ to $[0, 100]$.
We set the five focal lengths used for multi-focal capturing to $3.0$, $4.0$, $5.0$, $6.0$, $7.5$, and set the resolution of the rendered image to $512\times512$. When capturing the 3D mesh, we directly use the diffuse color of the texture without additional light source at the corresponding direction as the rendering result.
All experiments are conducted on an NVIDIA A100-80GB GPU. We will continuously review the latest methods and update our leaderboard.

To obtain an optimal mesh extraction, Marching Cubes is utilized for DreamFusion, LatentNeRF, ProlificDreamer and MVDream, while other methods employ DMTet; To retain quality without excessive UV unwrapping times, textures are extracted following mesh geometry simplification to a maximum of 40,000 faces.
For methods that yield 3D scenes with a diffusion latent radiance field representation rather than RGB, we also convert them into a latent texture map. Subsequently, we transform these into RGB textures using a latent decoder with a sliding window strategy to achieve anti-aliasing conversion.

\xhdr{Results}
Tab.~\ref{tb:results} reports the quality scores, alignment scores, and the average scores for each text-to-3D method on the three prompt sets in \ourmethod.
We also showcase some examples in Sec.~\ref{app:case}.
We summarize our key findings on current methods in the following.

\begin{table}[htb]
\centering
\caption{The average scores of text-to-3D methods on \ourmethod.}
\resizebox{0.75\linewidth}{!}{
\begin{tabular}{lcccc}
\toprule[2pt]
& Running Time \yr{$\downarrow$} & Quality \yr{$\uparrow$} & Alignment \yr{$\uparrow$} &\done Average \yr{$\uparrow$} \\
\midrule
\emph{Single Object} \\

Dreamfusion~\cite{poole2023dreamfusion}      & \texttt{30min}  & 24.9 & 24.0 &\done 24.4 \\
Magic3D~\cite{lin2023magic3d}          & \texttt{40min}  & 38.7 & 35.3 &\done 37.0 \\
LatentNeRF~\cite{metzer2022latentnerf}       & \texttt{65min}  & 34.2 & 32.0 &\done 33.1 \\
Fantasia3D~\cite{chen2023fantasia3d}       & \texttt{45min}  & 29.2 & 23.5 &\done 26.4 \\
SJC~\cite{wang2023score}              & \texttt{25min}  & 26.3 & 23.0 &\done 24.7 \\
ProlificDreamer~\cite{wang2023prolificdreamer}  & \texttt{240min} & 51.1 & 47.8 &\done 49.4 \\
MVDream~\cite{shi2023mvdream}          & \texttt{30min}  & 53.2 & 42.3 &\done 47.8 \\
DreamGaussian~\cite{tang2023dreamgaussian}    & \texttt{7min}   & 19.9 & 19.8 &\done 19.8 \\
GeoDream~\cite{ma2023geodream}         & \texttt{400min}  & 48.4 & 33.8     &\done 41.1 \\
RichDreamer~\cite{qiu2023richdreamer}      & \texttt{70min}  & 57.3 & 40.0 &\done 48.6 \\

\midrule
\multicolumn{3}{l}{\emph{Single Object with Surroundings}} \\

Dreamfusion~\cite{poole2023dreamfusion}      & \texttt{30min}  & 19.3 & 29.8 &\done 24.6 \\
Magic3D~\cite{lin2023magic3d}          & \texttt{40min}  & 29.8 & 41.0 &\done 35.4 \\
LatentNeRF~\cite{metzer2022latentnerf}       & \texttt{65min}  & 23.7 & 37.5 &\done 30.6 \\
Fantasia3D~\cite{chen2023fantasia3d}       & \texttt{45min}  & 21.9 & 32.0 &\done 27.0 \\
SJC~\cite{wang2023score}              & \texttt{25min}  & 17.3 & 22.3 &\done 19.8 \\
ProlificDreamer~\cite{wang2023prolificdreamer}  & \texttt{240min} & 42.5 & 47.0 &\done 44.8 \\
MVDream~\cite{shi2023mvdream}          & \texttt{30min}  & 36.3 & 48.5 &\done 42.4 \\
DreamGaussian~\cite{tang2023dreamgaussian}    & \texttt{7min}   & 10.4 & 17.8 &\done 14.1 \\
GeoDream~\cite{ma2023geodream}         &  \texttt{400min} & 35.2 & 34.5 &\done 34.9 \\
RichDreamer~\cite{qiu2023richdreamer}      & \texttt{70min}  & 43.9 & 42.3 &\done 43.1 \\

\midrule
\multicolumn{3}{l}{\emph{Multiple Objects}} \\

Dreamfusion~\cite{poole2023dreamfusion}      & \texttt{30min}  & 17.3 & 14.8 &\done 16.1 \\
Magic3D~\cite{lin2023magic3d}          & \texttt{40min}  & 26.6 & 24.8 &\done 25.7 \\
LatentNeRF~\cite{metzer2022latentnerf}       & \texttt{65min}  & 21.7 & 19.5 &\done 20.6 \\
Fantasia3D~\cite{chen2023fantasia3d}       & \texttt{45min}  & 22.7 & 14.3 &\done 18.5 \\
SJC~\cite{wang2023score}              & \texttt{25min}  & 17.7 & 5.8  &\done 11.7 \\
ProlificDreamer~\cite{wang2023prolificdreamer}  & \texttt{240min} & 45.7 & 25.8 &\done 35.8 \\
MVDream~\cite{shi2023mvdream}          & \texttt{30min}  & 39.0 & 28.5 &\done 33.8 \\
DreamGaussian~\cite{tang2023dreamgaussian}    & \texttt{7min}   & 12.3 & 9.5  &\done 10.9 \\
GeoDream~\cite{ma2023geodream}         & \texttt{400min}  & 34.3 &  16.5 &\done 25.4 \\
RichDreamer~\cite{qiu2023richdreamer}      & \texttt{70min}  & 34.8 & 22.0 &\done 28.4 \\

\bottomrule[2pt]
\end{tabular}
}
\label{tb:results}
\vspace{-10pt}
\end{table}

\xhdr{1. A simple combination of SDS and Stable Diffusion can cause density collapse}
For \textbf{DreamFusion}, the randomness inherent in Stable Diffusion's 2D guidance makes the direct application of Score Distillation Sampling (SDS) to supervise NeRF generation somewhat unstable. This can occasionally result in an inability to form effective density information during the optimization process, leading to failures that lower the overall score. 
\textbf{Magic3D} introduced an additional mesh refinement stage with high-resolution guidance, significantly improving the generated content's quality. However, the first stage still suffers from a high failure rate. \textbf{LatentNeRF} reduces this rate and boosts performance by optimizing in the latent domain rather than the RGB domain from the outset. Furthermore, compared to SDS, Score Jacobian Chaining (\textbf{SJC}) is less likely to lead to density collapse but tends to produce a large volume of sparse and floating density, making it difficult to extract high-quality meshes and reducing its practicality, as evidenced by our metrics.

\xhdr{2. The Efficiency of Current Text-to-3D Methods Requires Enhancement}
Current optimization approaches based on SDS typically necessitate thirty minutes or more, constraining further applications. \textbf{DreamGaussian} accelerates content generation by employing Gaussian splatting as scene representation instead of NeRF. However, the meshes extracted from Gaussians often suffer from poor quality, excessive smoothness, and disordered textures, leading to a need for overall performance improvement. Future research into Gaussian splatting techniques and text-to-3D framework with feed-forward design may hold the key to significantly boosting efficiency.

\xhdr{3. VSD achieves rich detail generation at the expense of efficiency and Janus faces}
Variational Score Distillation (VSD) proposed by \textbf{ProlificDreamer} optimizes the distribution of 3D scenes, showing significant benefits in generating detailed information across both single-object settings and complex prompts. Nonetheless, the use of VSD can sometimes lead to the introduction of extraneous details or geometric noise, adversely affecting human perception and BLIP captioning accuracy. This issue becomes more pronounced as the object count increases, leading to a decrease in alignment metrics. Additionally, VSD's modifications do not incorporate 3D or multi-view priors, allowing the persistence of the Janus problem in the generated outcomes. Employing appropriate geometry initialization may help mitigate these issues.

\xhdr{4. Geometry initialization techniques need improvement}
Implementing effective geometry initialization before optimization shows the potential to improve 3D content generation. While \textbf{Fantasia3D} excels in generating rich textures, its efficacy diminishes in complex scenes due to the less precise geometry it produces, as the supervision of geometry generation only through Stable Diffusion. \textbf{GeoDream}, on the other hand, generates a set of pseudo-multi-view images through models like MVDream and Zero123++\cite{shi2023zero123++} to initialize cost volumes, leading to more accurate geometry initialization. However, constrained by the performance of these models, inconsistencies may arise among the multi-view images, resulting in initialization failures (approximately one-fifth based on our benchmarks), highlighting the need for further performance improvement.

\xhdr{5. Leveraging multi-view diffusion models achieving commendable outcomes yet faces challenges with OOD problems}
The multi-view diffusion model introduced by \textbf{MVDream} demonstrates substantial quality improvements, as reflected in the scores. It also effectively solves the multi-view inconsistency problem that arises with other methods. \textbf{RichDreamer} further developed a multi-view diffusion model that incorporates depth, normal, and albedo information. 
Nevertheless, these methods encounter a limitation in more complex scenarios, where there is a tendency to omit certain elements of the object or environment, or to generate inaccurate colors in the outcomes. This may stem in part from the fact that multi-view diffusion was trained on Objaverse~\cite{deitke2023objaverse}, a 3D dataset comprised primarily of centered objects, which explains the struggle with certain out-of-distribution (OOD) cases.

\xhdr{Trends across different prompt sets}
As shown in Tab.~\ref{tb:results}, the overall performance is relatively good for the {\it Single Object} set, particularly for ProlificDreamer, Magic3D, and MVDream. However, when additional surrounding information is incorporated or when multiple objects are placed, the quality metrics for all methods experience varying degrees of degradation. 

In terms of alignment, some methods are able to reflect object information beyond the surroundings. This results in no significant decline in the {\it Single Object with Surroundings} set compared to the {\it Single Object} set. However, a noticeable decline is observed when the prompt set changes to {\it Multiple Objects}. This trend reflects the current issue with most works using Score Distillation Sampling (SDS) as guidance to supervise the generation of 3D scenes.
Specifically, SDS is relatively stable for single objects, but when the descriptions of the surroundings are added or when there are multiple objects in the scene, the appearance of the surroundings may have many possibilities after denoising steps.
There may be more possibilities for relative positions between multiple objects, leading to increased variability in the results generated by the diffusion model.
This in turn reduces the stability when supervising the generation of 3D scenes, resulting in a significant decline in the results.

In contrast, ProlificDreamer uses Variational Score Distillation (VSD) instead of SDS. By optimizing the distribution of the scene rather than directly optimizing the rendering results of the scene for 3D generation, ProlificDreamer demonstrates a clear advantage in complex scenarios. The multi-view diffusion guidance used by MVDream also shows superior performance on multi-object scenes.
MVDream tends to generate clearer and more favorable 3D shapes compared to ProlificDreamer, which can sometimes produce redundant densities. However, when dealing with out-of-distribution text prompts outside of the Objaverse dataset, MVDream sometimes struggles to fully capture all of the information from the text (e.g. generate a single object when the text refers to multiple ones).

\xhdr{Parallels and contrasts of the quality and alignment metrics}
It is worth noting that quality and alignment are not entirely correlated. Quality is more concerned with the geometry and subjective quality within a certain range, while alignment focuses on accurately restoring the information in the prompt. It is relatively sensitive to additional erroneous information, encouraging the generation of precise and clear 3D scenes.
For instance, the overall performance of Fantasia3D decreases markedly when generating multiple objects, as it fails to create precise 3D geometry, resulting in poor alignment compared to LatentNeRF. However, the quality of some generated objects is commendable with the obtained rich texture, making the overall quality higher than LatentNeRF.

ProlificDreamer typically generates more realistic textures, contributing to its superior quality. However, it sometimes generates a large amount of information not mentioned in the prompt, resulting in the possibility that the information described in the prompt only occupies a small part of the 3D generation results. Sometimes it only appears in the form of partial texture without significant geometry, which reduces its alignment index. Moreover, this characteristic is not what subsequent applications of text-to-3D want to see, further highlighting the importance of the alignment metric.

\xhdr{2D Guidance Analysis}
In Sec.~\ref{sec:2dguidance}, we investigate the effectiveness of 2D guidance from Stable Diffusion in generating 3D scenes by examining the correlation between the quality of 2D image generation and the quality of resulting 3D scenes. Results show that while Stable Diffusion produces high-quality 2D images, the ability of text-to-3D methods to utilize this guidance for accurate 3D scene generation is limited, reflected in generally low Spearman correlation between 2D image quality and 3D scene quality. The findings highlight that the main challenges in text-to-3D generation are learning 3D structures from 2D guidance and ensuring view consistency.

%% file: sec/5_conclusion.tex
\section{Conclusion}

In this work, we present \ourmethod, the first comprehensive benchmark for evaluating text-to-3D generation methods. \ourmethod\ serves as a rich testbed as it provides diverse prompt suites, and supports fully automatic evaluation by incorporating our proposed multi-view quality and alignment metrics that closely correlate with human judgments. 
We carefully benchmark 10 prevalent text-to-3D methods on \ourmethod, and highlight
a number of common and specific problems with current methods.

\section{Discussion}

\xhdr{Size of Data} Unlike existing text-to-image methods that enable efficient generation, the current text-to-3D techniques are considerably slower, requiring a minimum of half an hour and potentially several hours for a single prompt. This makes it hard to test with larger sets of prompts.

\xhdr{Indirect Evaluation} Given the absence of an effective evaluation method that directly aligns the generated 3D scenes with human evaluation, there is an inevitable loss of information during the 3D to 2D rendering process, even with the efficacy of our multi-view capturing and processing scheme in evaluating geometry and other information.
Likewise, no 3D captioning framework matches the performance of BLIP in 2D image captioning. 
While our multi-view captioning and merging strategy typically generates accurate 3D captions, the merging process does not always yield flawless results.

%% file: sec/X_suppl.tex
\clearpage
\setcounter{page}{1}

\section{Example Prompts}
\subsection{Question Generation}

\xhdr{Single Object}
\begin{quote}
\begin{em}
Please describe 20 objects' appearance for me in brief words, without background. Please make sure that the object you provided has enough diversity, and that the format is similar to my example. Here is an example: ``A pig wearing a backpack''.
\end{em}
\end{quote}

\xhdr{Single Object with Surroundings}
\begin{quote}
\begin{em}
Please describe 20 objects for me in brief words. Please make sure that the object you provided has enough diversity, and that the format is similar to my example. Here is an example: ``A black metal bicycle leaning against a brick wall''.
\end{em}
\end{quote}

\xhdr{Multiple Objects}
\begin{quote}
\begin{em}
Please describe 20 different scenes for me in brief words, each scene contains multiple objects. Do not describe the environment. Please make sure that the scenes you provided have enough diversity, and the format similar to my example. Here are examples: ``A child with a red shirt is playing with a dog'', or ``Two coffee cups stand on the table''.
\end{em}
\end{quote}

\subsection{Multiple Caption Merging}
\label{ssec:caption_merging}
\begin{quote}
\begin{em}
Given a set of descriptions about the same 3D object, distill these descriptions into one concise caption. The descriptions are as follows:

view1: ... \\
view2: ... \\
... \\
view\{N\}: ...

Avoid describing background, surface, and posture. The caption should be:
\end{em}
\end{quote}

\subsection{LLM Likert Scale Scoring}
\begin{quote}
\begin{em}
You are an assessment expert responsible for prompt-prediction pairs. Your task is to score the prediction according to the following requirements:

1. Evaluate the recall, or how well the prediction covers the information in the prompt. if the prediction contains information that does not appear in the prompt, it should not be considered as bad. 

2. If the prediction contains correct information about color or features in the prompt, you should also consider raising your score.

3. Assign a score between 1 and 5, with 5 being the highest. Do not provide a complete answer; give the score in the format: 3

Prompt: ...

Prediction: ...
\end{em}
\end{quote}

\section{Experimental Details}

\subsection{Metric Evaluation}
\label{app:metric}
For the evaluation of metrics, we randomly select 30\% of the prompts from each prompt set, along with their corresponding 3D mesh generated by the text-to-3D method. This results in a total of 630 samples. We request human annotators to carefully check the mesh in an interactive 3D viewer and score the responses on a scale of 1-5, based on their 3D quality and alignment. Below, we provide the annotation instructions:

\begin{mdframed}
1. Scoring is based on two dimensions: quality (which assesses the subjective quality of the 3D generation) and alignment (which evaluates how well the generated content covers the original prompt). These two dimensions are scored on a scale of 1 to 5, with 1 being the lowest and 5 the highest. \\
2. Please drag each generated mesh to our specified 3D viewer. After carefully examining the mesh from various angles, assign your score based on the above two dimensions.
\end{mdframed}

\subsection{Capture Viewpoint Selection}
\label{app:ico}
\begin{figure}[t]
\centering
\includegraphics[width=0.7\linewidth]{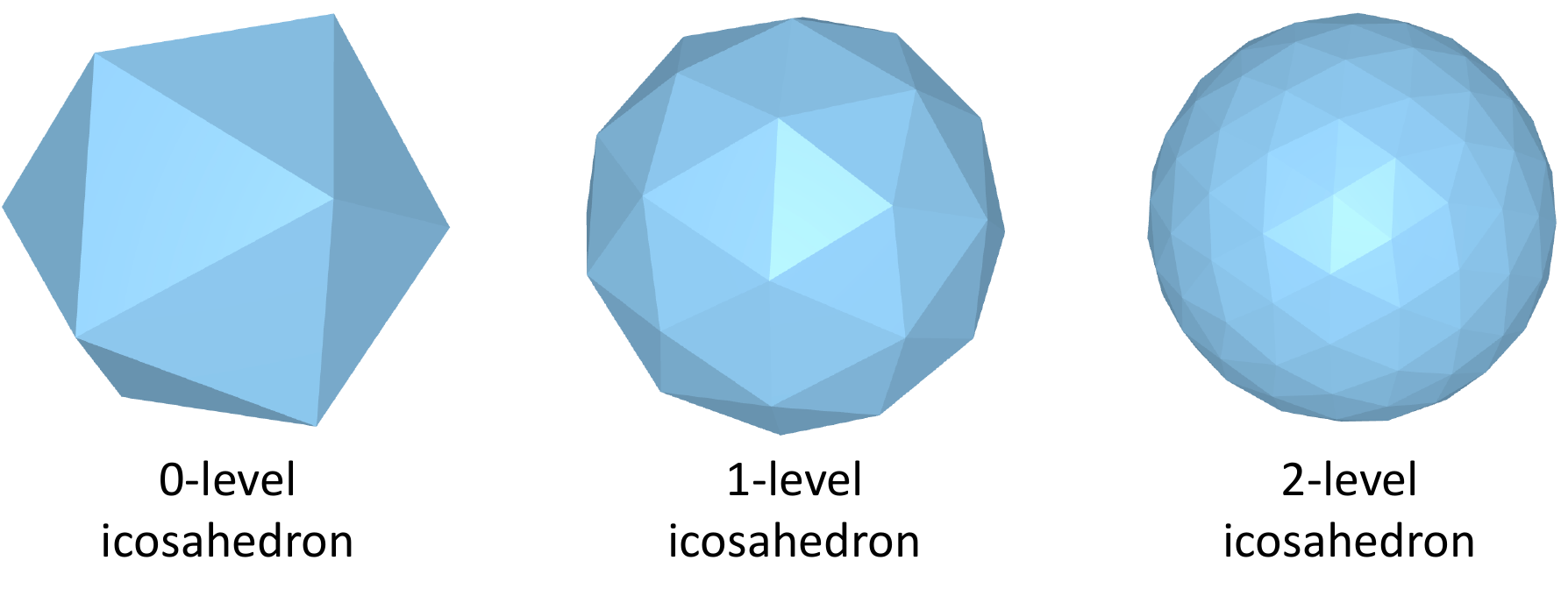}
\caption{ Schema of icosahedrons with different levels. }
\label{fig:ico}
\end{figure}

In order to uniformly select the capturing location of the 3D mesh, we construct the icosahedron and use its vertices as the location for the captures. The vertex coordinates of a level-0 unit icosahedron are computed as follows:

\begin{align}
    V^{(0)}=\sqrt{1+\phi^2}\cdot
    \left|\begin{array}{ccc}\phi & 1 & 0 \\ -\phi & 1 & 0 \\ \phi & -1 & 0 \\ -\phi & -1 & 0 \\ 1 & 0 & \phi \\ 1 & 0 & -\phi \\ -1 & 0 & \phi \\ -1 & 0 & -\phi \\ 0 & \phi & 1 \\ 0 & -\phi & 1 \\ 0 & \phi & -1 \\ 0 & -\phi & -1\end{array}\right|,
\end{align}
\noindent
where

\begin{align}
    \phi = \frac{1+\sqrt{5}}{2},
\end{align}
\noindent
and there is an edge between every two points with a distance of $2/\sqrt{1+\phi^2}$, resulting in 12 vertices, 30 edges, and 20 triangle faces.

A level-$K$ unit icosahedron can be obtained recursively by adding an extra vertice on every edge of a level-($K$-1) unit icosahedron and adding an edge between every two new vertexes with a triangle face of the level-($K$-1) unit icosahedron, then scaling every new vertex's coordinate to a length of 1. A demonstration of different level icosahedrons is shown in Fig.~\ref{fig:ico}.

\subsection{Capturing Poses Derivation}
Since many text image scoring models are sensitive to rotation, we need to make sure that the angle of the shot is as free as possible from 2D rotation around the look-at vector. We ensure this constraint with the following procedure:

Given the location v of the shot, we can get the look-at vector as follows:

\begin{align}
    \textbf{l}=-\frac{\textbf{v}}{||\textbf{v}||}.
\end{align}

Then, we acquire the horizontal vector $\textbf{r}$ of the camera plane by

\begin{align}
    \textbf{r} = \frac{\textbf{u} \times \textbf{l}}{||\textbf{u} \times \textbf{l}||},
\end{align}
\noindent
where $\textbf{u}$ is the unit vector parallel with the positive direction of the vertical axis. The up vector of the camera plane can be calculated by

\begin{align}
    \textbf{u}' =  \textbf{l} \times \textbf{r}.
\end{align}
\noindent
Finally, the camera matrix \textbf{P} is formed with

\begin{align}
    \textbf{P} = [-\textbf{r}\quad \textbf{u}'\quad \textbf{l}\quad \textbf{v}].
\end{align}

\section{Design Choices for Regional Convolution}
\label{sec:conv}
The general form of regional convolution can be formed as:

\begin{equation}
s_{i}^{(t+1)} = \frac{1}{w|N(i)|+1} \left( s_{i}^{(t)} + w\sum_{j \in N(i)} s_{j}^{(t)} \right),
\end{equation}

To further explore the use of convolution kernels and the impact of the receptive field on the quality evaluation metric, we conduct experiments on correlations with human annotations, where the convolution weights \(w\) are varied from 0 to 2, and different times of convolutions are applied.

\begin{figure}[htb]
\centering
\includegraphics[width=\linewidth]{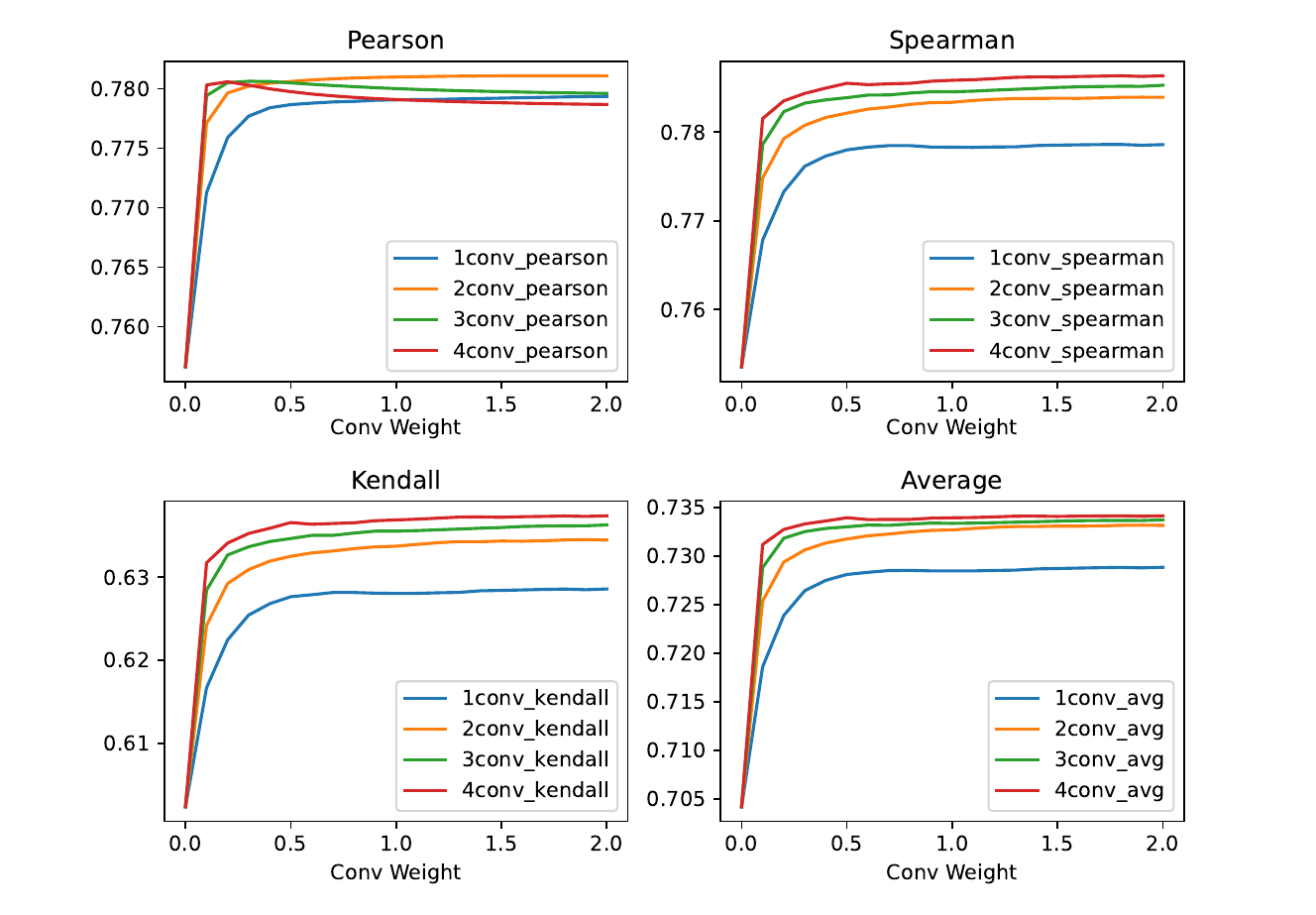}
\caption{Correlation variations on different weights and times of convolutions.}
\label{fig:convs}
\end{figure}

We observe that as the number of convolutions and the weights of neighbors in convolutions increases, both the Spearman and Kendall correlation coefficients consistently rise. However, the Pearson correlation coefficient exhibits a more complex trend. Specifically, it decreases when the number of convolutions and the weights of neighboring convolutions are excessively high. This phenomenon could be attributed to a smoother convolution operation with a larger reception field, which takes a more overall sense, is more beneficial for preserving the order of evaluation metrics. However, excessive smoothing can overlook finer details such as texture quality, leading to a non-uniform compression of scores and a reduction in the linearity of the quality evaluation metric. Considering these factors, we ultimately selected a weight \(w=1\) and applied three times of convolutions.

\section{2D Guidance Analysis}
\label{sec:2dguidance}

The majority of current text-to-3D methods utilize 2D priors associated with Stable Diffusion~\cite{rombach2022high} for the generation of 3D scenes.
To delve deeper into the effectiveness of 2D guidance and the capabilities of current text-to-3D methods in utilizing this guidance for 3D generation, we explore the correlation between the quality of 2D image generation produced by the diffusion model and the resulting quality of the 3D generation. For each prompt in \ourmethod, we apply the Stable Diffusion backbone of each method for text-to-image generation.
Notably, the text-to-3D methods utilize view-dependent prompting in conjunction with 2D guidance from the diffusion model during the generation process. Descriptions of viewing angles (e.g. front view, side view) are added at the end of the prompt. Given that the range and granularity of viewing descriptions in view-dependent prompting vary across different text-to-3D methods, we directly use the original prompt without view-dependent prompting in the text-to-image generation.
We then compute the single-view quality metric on the generated 2D image. Finally,
we compute the correlation between the single-view quality metric of the generated 2D image and the quality metric (Multi-view capturing with ImageReward) result of the generated 3D scene.

\begin{table}[htb]
\centering
\caption{The Spearman's $\rho$ correlation between the text-to-3D methods' generation qualities and the diffusion models' 2D image generation qualities, averaged over all prompts.}
\resizebox{0.8\linewidth}{!}{
\begin{tabular}{lccc}
\toprule[2pt]
& Single Obj. & \quad Single Obj. with Surr. \quad & Multi Obj. \\
\midrule

Dreamfusion      & 0.211 & 0.184 & 0.045 \\
Magic3D          & 0.229 & 0.158 & 0.059 \\
LatentNeRF       & 0.290 & 0.191 & 0.050 \\
Fantasia3D       & 0.159 & 0.153 & 0.006 \\
SJC              & 0.228 & 0.159 & 0.040 \\
ProlificDreamer  & 0.357 & 0.272 & 0.147 \\
MVDream          & 0.421 & 0.340 & 0.474 \\
DreamGaussian    & 0.206 & 0.132 & 0.156 \\
GeoDream         & 0.330 & 0.228 & 0.086 \\
Richdreamer      & 0.407 & 0.347 & 0.252 \\

\bottomrule[2pt]
\end{tabular}
}
\label{tb:2dcorr}
\end{table}

Tab.~\ref{tb:2dcorr} displays the Spearman correlation between the text-to-image scores for the 2D guidance and the final text-to-3D scores. It can be observed that that all correlations are relatively low, and there are two overall trends: 1) methods demonstrating better performance in text-to-3D also have higher correlation coefficients; and 2) when using different prompt sets, the correlation coefficient also follows the trend of {\it Single Object} greater than {\it Single Object with Surroundings}, and the latter greater than {\it Multiple Objects}. 
We attribute these outcomes to the fact that Stable Diffusion can generate satisfactory 2D images most of the time, even for complex prompts.
However, 2D guidance may not be effectively used by text-to-3D methods --- they may fail to generate accurate 3D scenes even though the 2D images are acceptable, leading to a low text-to-3D score while high text-to-image score.
In addition, the 2D guidance may not be view-consistent, which does not significantly affect the text-to-image scores but can indeed lead to poorer quality in the final 3D generation.
Superior methods like ProlificDreamer can better utilize 2D images to form a 3D scene, as suggested by its higher correlation, and as a result, can generate higher quality 3D scenes.

The retrained multi-view diffusion model by MVDream (also leveraged by RichDreamer) provides effective guidance for 3D generation, as evidenced by the highest correlation results. This highlights the capabilities of the retrained diffusion model. However, the retrained diffusion itself exhibits a degree of degradation in its generation capabilities, especially in scenarios involving surrounding information and multiple objects. This is reflected in the lower average scores of 2D image generation with MVDream's diffusion model compared to Stable Diffusion (e.g. 32.9 vs 44.0 on the multi-object set). While the retrained diffusion model is useful for guiding 3D generation, there is still room for improvement in diffusion modeling for 3D tasks.

These observations suggest that the current bottleneck of text-to-3D lies in the process of learning 3D from 2D guidance and the view consistency of 2D guidance, rather than the generative capability of Stable Diffusion itself.

\section{More Case Studies}
\label{app:case}

\subsection{Single-view vs. Multi-view Capturing}

\begin{figure}[t]
\centering
\includegraphics[width=1.0\linewidth]{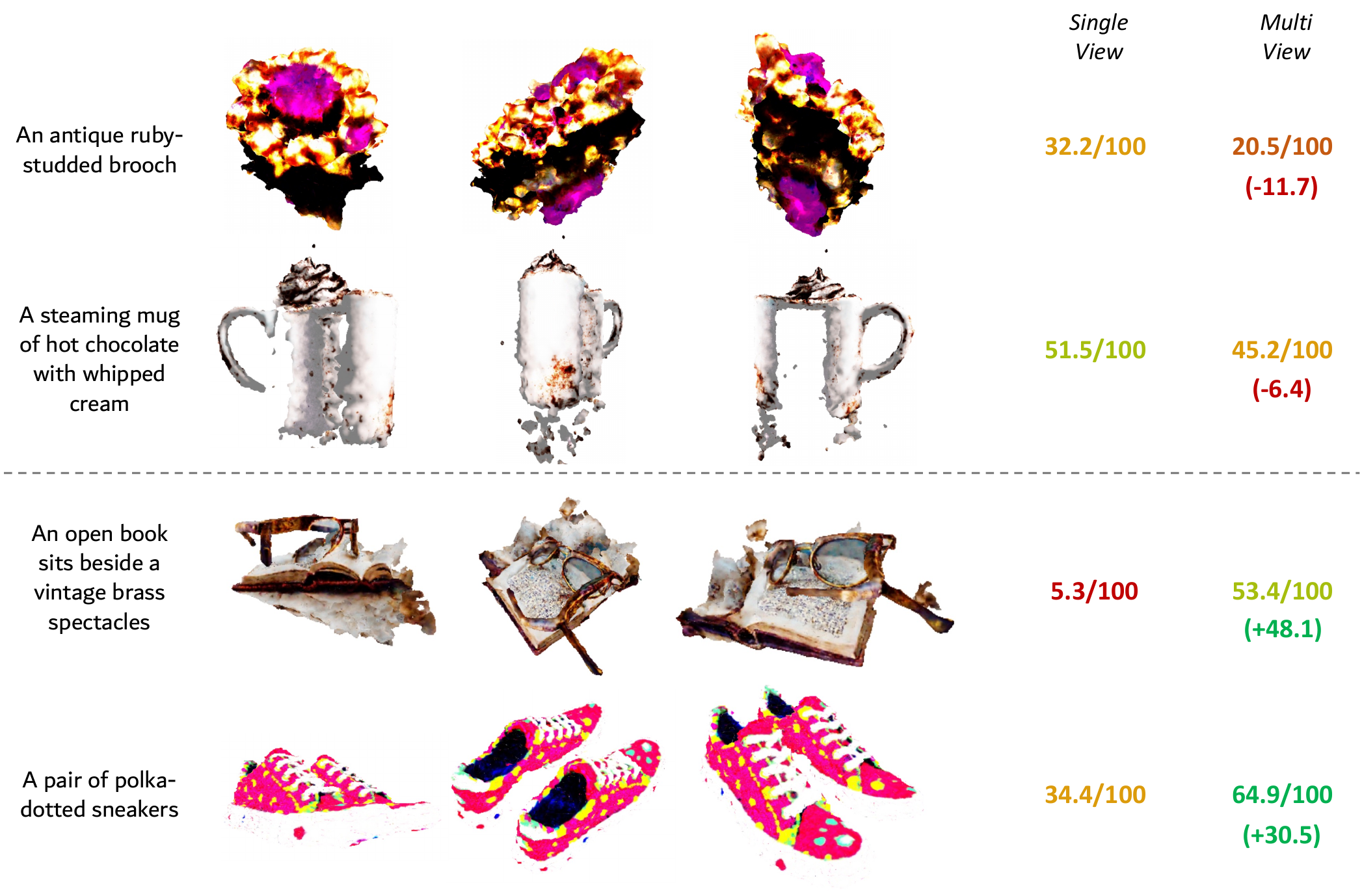}
\caption{Comparisons of the scoring between single-view capturing and our multi-view capturing scheme. The first image column denotes the single front view, and the other two image columns are captured from other viewpoints.}
\label{fig:single}
\end{figure}

We further illustrate through a case study that adhering to the previous method and only capturing single-view images does not yield satisfactory evaluations. As depicted in Fig.~\ref{fig:single}, the first two examples demonstrate good subjective quality in the front view. However, their geometries are relatively poor, and there are noticeable residuals or artifacts when they are converted to other viewpoints. These can be identified with our multi-view capturing mechanism, which subsequently adjusts the scores accordingly. In the next two examples, the front view is partially obscured, which fails to fully represent the subjective quality of the generated objects. Our multi-view capturing mechanism can detect this and improve their scores accordingly.

\subsection{More results}
\begin{figure*}[h]
\centering
\includegraphics[width=0.9\linewidth]{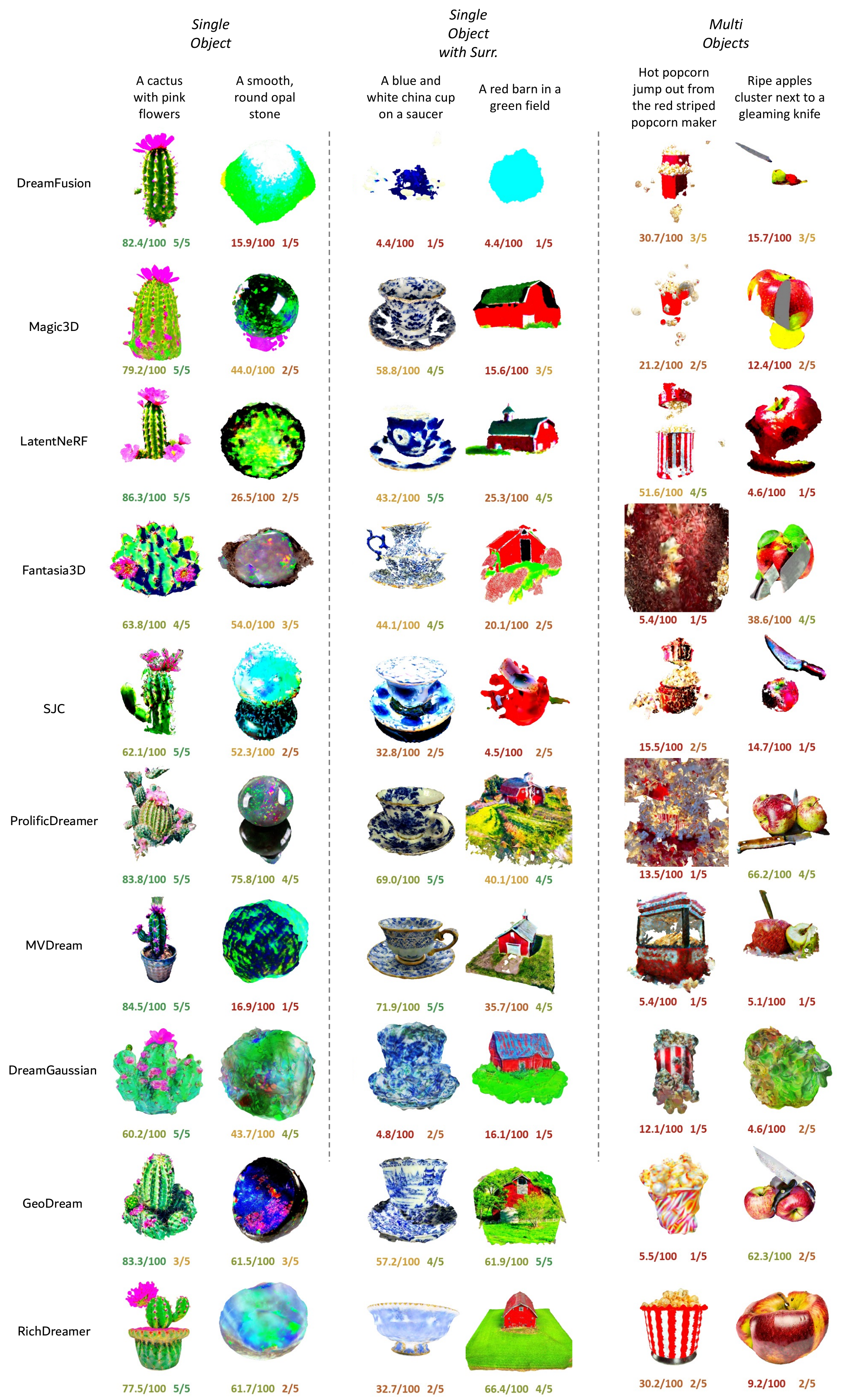}
\caption{Visualizations of text-to-3D generation results. The two scores denote quality and alignment, respectively.}
\label{fig:result}
\end{figure*}

\begin{figure*}[t]
\centering
\includegraphics[width=1\linewidth]{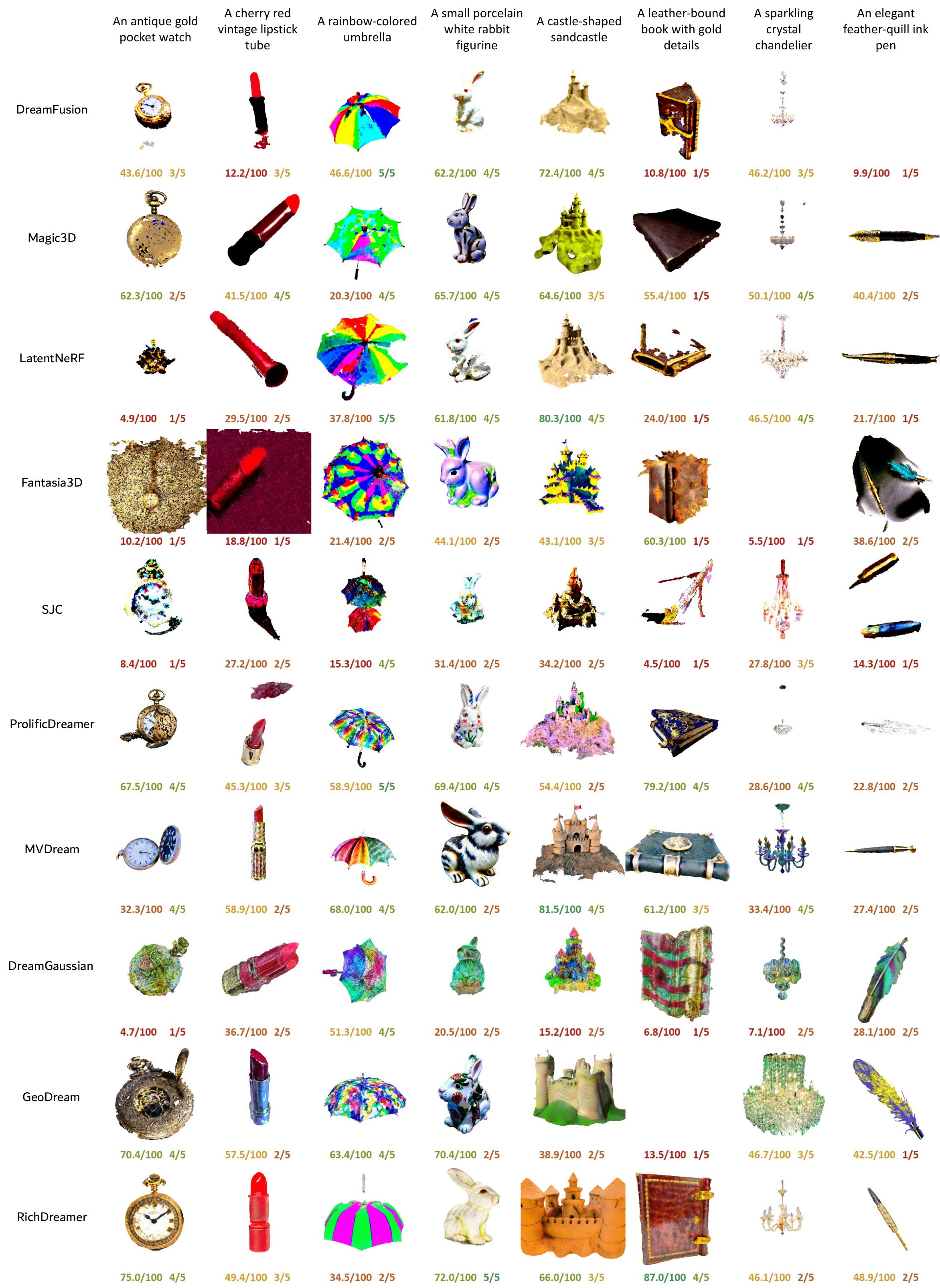}
\caption{ More results of our test prompts, including generations and evaluations of different text-to-3D methods (\#1). }
\label{fig:res1}
\end{figure*}

\begin{figure*}[t]
\centering
\includegraphics[width=1\linewidth]{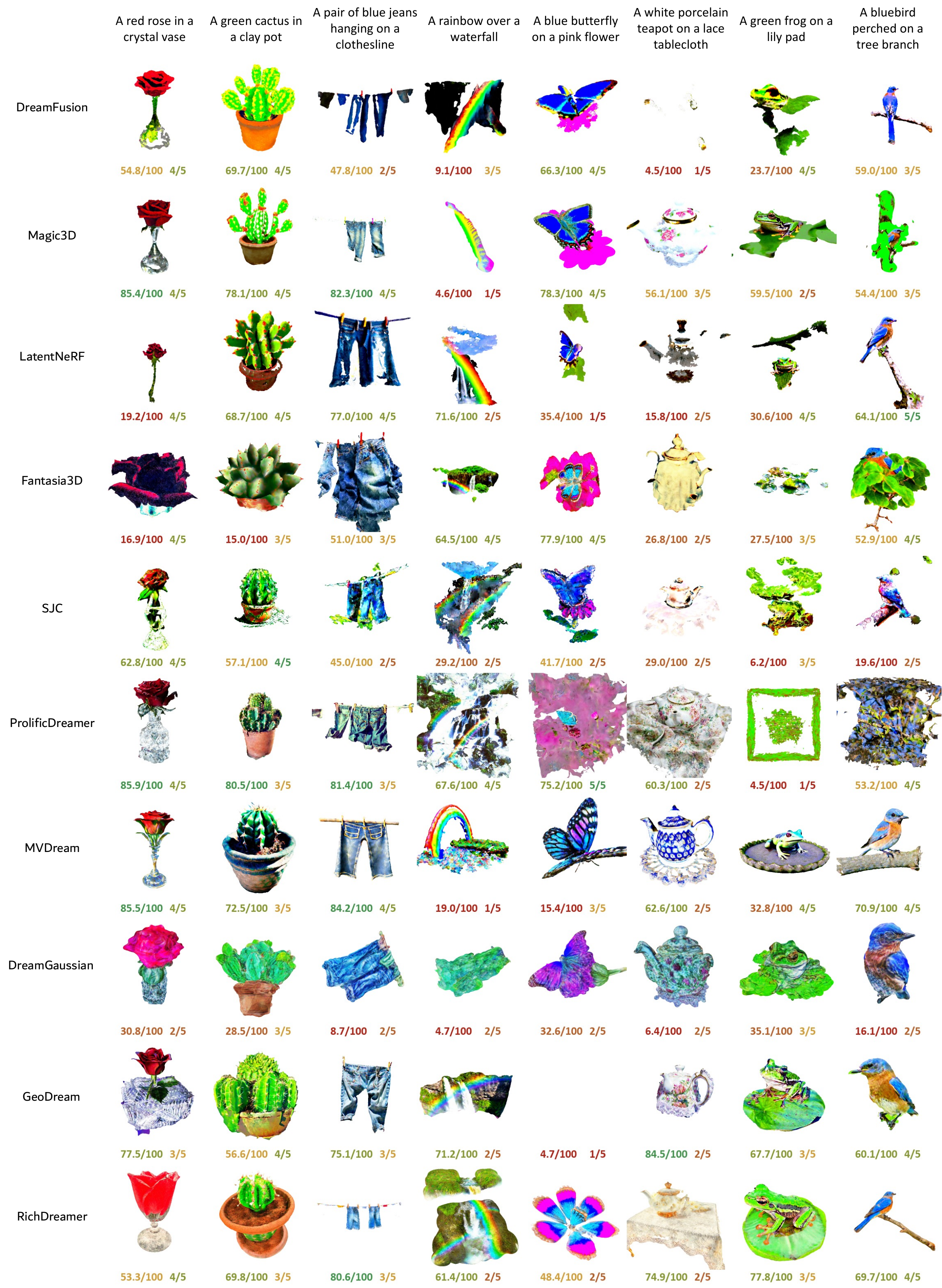}
\caption{ More results of our test prompts, including generations and evaluations of different text-to-3D methods (\#2). }
\label{fig:res2}
\end{figure*}

\begin{figure*}[t]
\centering
\includegraphics[width=1\linewidth]{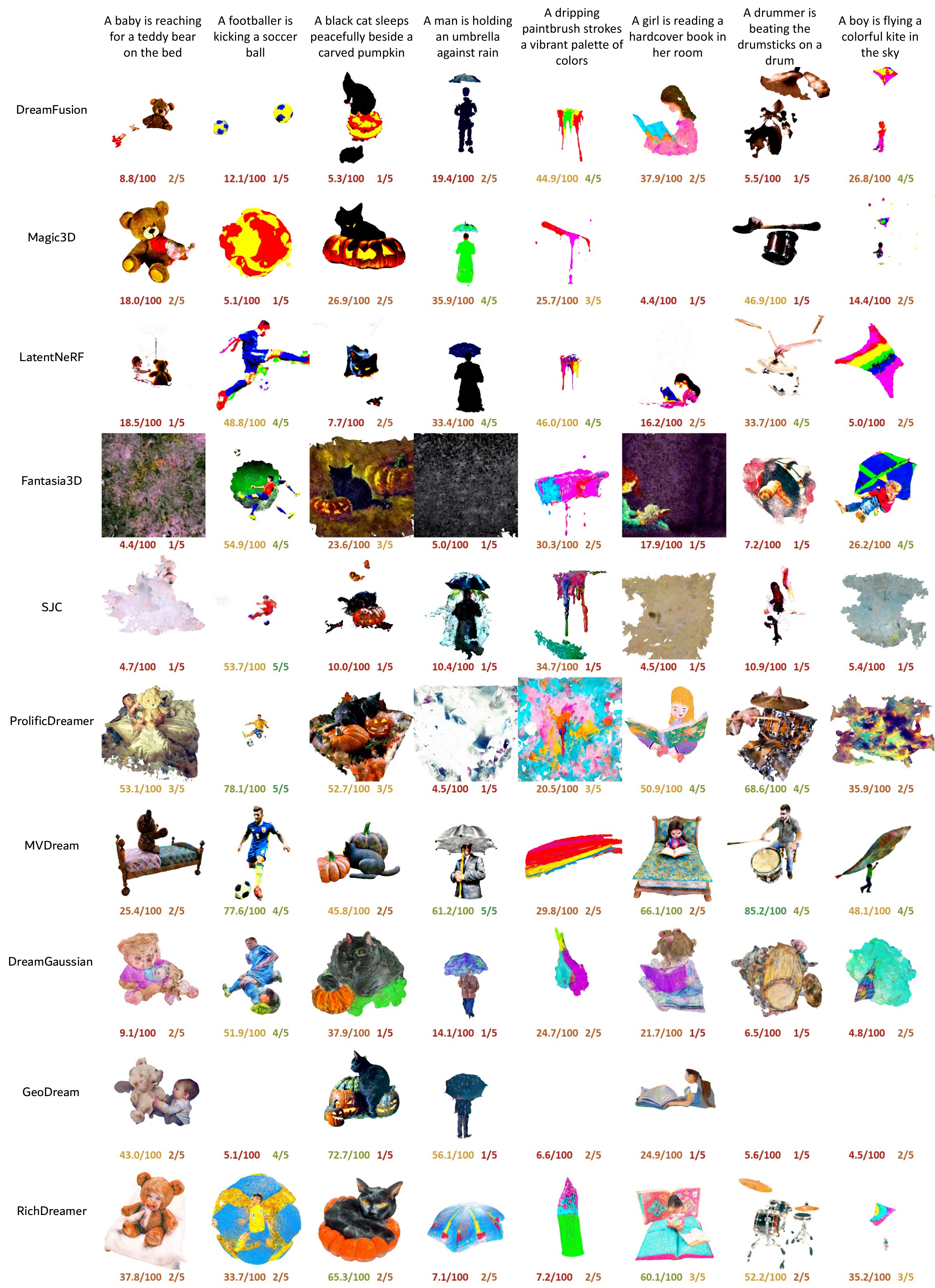}
\caption{ More results of our test prompts, including generations and evaluations of different text-to-3D methods (\#3). }
\label{fig:res3}
\end{figure*}

We provide case studies of test prompts with generations and evaluations of different text-to-3D methods in Figs.~\ref{fig:result},~\ref{fig:res1},~\ref{fig:res2},~\ref{fig:res3}.

%% file: main.bbl
\begin{thebibliography}{10}
\providecommand{\url}[1]{\texttt{#1}}
\providecommand{\urlprefix}{URL }
\providecommand{\doi}[1]{https://doi.org/#1}

\bibitem{bai2023benchmarking}
Bai, Y., Ying, J., Cao, Y., Lv, X., He, Y., Wang, X., Yu, J., Zeng, K., Xiao, Y., Lyu, H., et~al.: Benchmarking foundation models with language-model-as-an-examiner. arXiv preprint arXiv:2306.04181  (2023)

\bibitem{bakr2023hrs}
Bakr, E.M., Sun, P., Shen, X., Khan, F.F., Li, L.E., Elhoseiny, M.: Hrs-bench: Holistic, reliable and scalable benchmark for text-to-image models. arXiv preprint arXiv:2304.05390  (2023)

\bibitem{brooks2023instructpix2pix}
Brooks, T., Holynski, A., Efros, A.A.: Instructpix2pix: Learning to follow image editing instructions. In: Proceedings of the IEEE/CVF Conference on Computer Vision and Pattern Recognition. pp. 18392--18402 (2023)

\bibitem{chen2023fantasia3d}
Chen, R., Chen, Y., Jiao, N., Jia, K.: Fantasia3d: Disentangling geometry and appearance for high-quality text-to-3d content creation. arXiv preprint arXiv:2303.13873  (2023)

\bibitem{cho2023dalleval}
Cho, J., Zala, A., Bansal, M.: Dall-eval: Probing the reasoning skills and social biases of text-to-image generation models (2023)

\bibitem{deitke2023objaverse}
Deitke, M., Schwenk, D., Salvador, J., Weihs, L., Michel, O., VanderBilt, E., Schmidt, L., Ehsani, K., Kembhavi, A., Farhadi, A.: Objaverse: A universe of annotated 3d objects. In: Proceedings of the IEEE/CVF Conference on Computer Vision and Pattern Recognition. pp. 13142--13153 (2023)

\bibitem{threestudio2023}
Guo, Y.C., Liu, Y.T., Shao, R., Laforte, C., Voleti, V., Luo, G., Chen, C.H., Zou, Z.X., Wang, C., Cao, Y.P., Zhang, S.H.: threestudio: A unified framework for 3d content generation. \url{https://github.com/threestudio-project/threestudio} (2023)

\bibitem{he2023mmpi}
He, Y., Wang, P., Hu, Y., Zhao, W., Yi, R., Liu, Y.J., Wang, W.: Mmpi: a flexible radiance field representation by multiple multi-plane images blending (2023)

\bibitem{hertz2022prompt}
Hertz, A., Mokady, R., Tenenbaum, J., Aberman, K., Pritch, Y., Cohen-Or, D.: Prompt-to-prompt image editing with cross attention control. arXiv preprint arXiv:2208.01626  (2022)

\bibitem{ho2020denoising}
Ho, J., Jain, A., Abbeel, P.: Denoising diffusion probabilistic models. Advances in neural information processing systems  \textbf{33},  6840--6851 (2020)

\bibitem{hong2023debiasing}
Hong, S., Ahn, D., Kim, S.: Debiasing scores and prompts of 2d diffusion for robust text-to-3d generation. arXiv preprint arXiv:2303.15413  (2023)

\bibitem{jain2022zero}
Jain, A., Mildenhall, B., Barron, J.T., Abbeel, P., Poole, B.: Zero-shot text-guided object generation with dream fields. In: Proceedings of the IEEE/CVF Conference on Computer Vision and Pattern Recognition. pp. 867--876 (2022)

\bibitem{lee2024holistic}
Lee, T., Yasunaga, M., Meng, C., Mai, Y., Park, J.S., Gupta, A., Zhang, Y., Narayanan, D., Teufel, H., Bellagente, M., et~al.: Holistic evaluation of text-to-image models. Advances in Neural Information Processing Systems  \textbf{36} (2024)

\bibitem{li2023instant3d}
Li, J., Tan, H., Zhang, K., Xu, Z., Luan, F., Xu, Y., Hong, Y., Sunkavalli, K., Shakhnarovich, G., Bi, S.: Instant3d: Fast text-to-3d with sparse-view generation and large reconstruction model. arXiv preprint arXiv:2311.06214  (2023)

\bibitem{li2022blip}
Li, J., Li, D., Xiong, C., Hoi, S.: Blip: Bootstrapping language-image pre-training for unified vision-language understanding and generation. In: International Conference on Machine Learning. pp. 12888--12900. PMLR (2022)

\bibitem{lin2023magic3d}
Lin, C.H., Gao, J., Tang, L., Takikawa, T., Zeng, X., Huang, X., Kreis, K., Fidler, S., Liu, M.Y., Lin, T.Y.: Magic3d: High-resolution text-to-3d content creation. In: Proceedings of the IEEE/CVF Conference on Computer Vision and Pattern Recognition. pp. 300--309 (2023)

\bibitem{lin2004rouge}
Lin, C.Y.: Rouge: A package for automatic evaluation of summaries. In: Text summarization branches out. pp. 74--81 (2004)

\bibitem{lin2014microsoft}
Lin, T.Y., Maire, M., Belongie, S., Hays, J., Perona, P., Ramanan, D., Doll{\'a}r, P., Zitnick, C.L.: Microsoft coco: Common objects in context. In: Computer Vision--ECCV 2014: 13th European Conference, Zurich, Switzerland, September 6-12, 2014, Proceedings, Part V 13. pp. 740--755. Springer (2014)

\bibitem{lorensen1998marching}
Lorensen, W.E., Cline, H.E.: Marching cubes: A high resolution 3d surface construction algorithm. In: Seminal graphics: pioneering efforts that shaped the field, pp. 347--353 (1998)

\bibitem{luo2023scalable}
Luo, T., Rockwell, C., Lee, H., Johnson, J.: Scalable 3d captioning with pretrained models. arXiv preprint arXiv:2306.07279  (2023)

\bibitem{ma2023geodream}
Ma, B., Deng, H., Zhou, J., Liu, Y.S., Huang, T., Wang, X.: Geodream: Disentangling 2d and geometric priors for high-fidelity and consistent 3d generation. arXiv preprint arXiv:2311.17971  (2023)

\bibitem{metzer2022latentnerf}
Metzer, G., Richardson, E., Patashnik, O., Giryes, R., Cohen-Or, D.: Latent-nerf for shape-guided generation of 3d shapes and textures (2022)

\bibitem{mildenhall2021nerf}
Mildenhall, B., Srinivasan, P.P., Tancik, M., Barron, J.T., Ramamoorthi, R., Ng, R.: Nerf: Representing scenes as neural radiance fields for view synthesis. Communications of the ACM  \textbf{65}(1),  99--106 (2021)

\bibitem{mohammad2022clip}
Mohammad~Khalid, N., Xie, T., Belilovsky, E., Popa, T.: Clip-mesh: Generating textured meshes from text using pretrained image-text models. In: SIGGRAPH Asia 2022 conference papers. pp.~1--8 (2022)

\bibitem{openai2023gpt4}
OpenAI: Gpt-4 technical report. arXiv preprint arXiv:2303.08774  (2023)

\bibitem{poole2022dreamfusion}
Poole, B., Jain, A., Barron, J.T., Mildenhall, B.: Dreamfusion: Text-to-3d using 2d diffusion. arXiv preprint arXiv:2209.14988  (2022)

\bibitem{poole2023dreamfusion}
Poole, B., Jain, A., Barron, J.T., Mildenhall, B.: Dreamfusion: Text-to-3d using 2d diffusion. In: The Eleventh International Conference on Learning Representations (2023)

\bibitem{qian2023magic123}
Qian, G., Mai, J., Hamdi, A., Ren, J., Siarohin, A., Li, B., Lee, H.Y., Skorokhodov, I., Wonka, P., Tulyakov, S., et~al.: Magic123: One image to high-quality 3d object generation using both 2d and 3d diffusion priors. arXiv preprint arXiv:2306.17843  (2023)

\bibitem{qiu2023richdreamer}
Qiu, L., Chen, G., Gu, X., Zuo, Q., Xu, M., Wu, Y., Yuan, W., Dong, Z., Bo, L., Han, X.: Richdreamer: A generalizable normal-depth diffusion model for detail richness in text-to-3d. arXiv preprint arXiv:2311.16918  (2023)

\bibitem{radford2021learning}
Radford, A., Kim, J.W., Hallacy, C., Ramesh, A., Goh, G., Agarwal, S., Sastry, G., Askell, A., Mishkin, P., Clark, J., et~al.: Learning transferable visual models from natural language supervision. In: International conference on machine learning. pp. 8748--8763. PMLR (2021)

\bibitem{rombach2022high}
Rombach, R., Blattmann, A., Lorenz, D., Esser, P., Ommer, B.: High-resolution image synthesis with latent diffusion models. In: Proceedings of the IEEE/CVF conference on computer vision and pattern recognition. pp. 10684--10695 (2022)

\bibitem{stablediffusion}
Rombach, R., Blattmann, A., Lorenz, D., Esser, P., Ommer, B.: High-resolution image synthesis with latent diffusion models. In: Proceedings of the IEEE/CVF conference on computer vision and pattern recognition. pp. 10684--10695 (2022)

\bibitem{saharia2022photorealistic}
Saharia, C., Chan, W., Saxena, S., Li, L., Whang, J., Denton, E., Ghasemipour, S.K.S., Ayan, B.K., Mahdavi, S.S., Lopes, R.G., Salimans, T., Ho, J., Fleet, D.J., Norouzi, M.: Photorealistic text-to-image diffusion models with deep language understanding (2022)

\bibitem{imagen}
Saharia, C., Chan, W., Saxena, S., Li, L., Whang, J., Denton, E.L., Ghasemipour, K., Gontijo~Lopes, R., Karagol~Ayan, B., Salimans, T., et~al.: Photorealistic text-to-image diffusion models with deep language understanding. Advances in Neural Information Processing Systems  \textbf{35},  36479--36494 (2022)

\bibitem{schuhmann2022laion}
Schuhmann, C., Beaumont, R., Vencu, R., Gordon, C., Wightman, R., Cherti, M., Coombes, T., Katta, A., Mullis, C., Wortsman, M., et~al.: Laion-5b: An open large-scale dataset for training next generation image-text models. Advances in Neural Information Processing Systems  \textbf{35},  25278--25294 (2022)

\bibitem{seo2023ditto}
Seo, H., Kim, H., Kim, G., Chun, S.Y.: Ditto-nerf: Diffusion-based iterative text to omni-directional 3d model. arXiv preprint arXiv:2304.02827  (2023)

\bibitem{seo2023let}
Seo, J., Jang, W., Kwak, M.S., Ko, J., Kim, H., Kim, J., Kim, J.H., Lee, J., Kim, S.: Let 2d diffusion model know 3d-consistency for robust text-to-3d generation. arXiv preprint arXiv:2303.07937  (2023)

\bibitem{shen2021deep}
Shen, T., Gao, J., Yin, K., Liu, M.Y., Fidler, S.: Deep marching tetrahedra: a hybrid representation for high-resolution 3d shape synthesis. Advances in Neural Information Processing Systems  \textbf{34},  6087--6101 (2021)

\bibitem{shi2023zero123++}
Shi, R., Chen, H., Zhang, Z., Liu, M., Xu, C., Wei, X., Chen, L., Zeng, C., Su, H.: Zero123++: a single image to consistent multi-view diffusion base model. arXiv preprint arXiv:2310.15110  (2023)

\bibitem{shi2023mvdream}
Shi, Y., Wang, P., Ye, J., Long, M., Li, K., Yang, X.: Mvdream: Multi-view diffusion for 3d generation. arXiv preprint arXiv:2308.16512  (2023)

\bibitem{tang2023dreamgaussian}
Tang, J., Ren, J., Zhou, H., Liu, Z., Zeng, G.: Dreamgaussian: Generative gaussian splatting for efficient 3d content creation. arXiv preprint arXiv:2309.16653  (2023)

\bibitem{tsalicoglou2023textmesh}
Tsalicoglou, C., Manhardt, F., Tonioni, A., Niemeyer, M., Tombari, F.: Textmesh: Generation of realistic 3d meshes from text prompts. arXiv preprint arXiv:2304.12439  (2023)

\bibitem{wang2023score}
Wang, H., Du, X., Li, J., Yeh, R.A., Shakhnarovich, G.: Score jacobian chaining: Lifting pretrained 2d diffusion models for 3d generation. In: Proceedings of the IEEE/CVF Conference on Computer Vision and Pattern Recognition. pp. 12619--12629 (2023)

\bibitem{wang2023prolificdreamer}
Wang, Z., Lu, C., Wang, Y., Bao, F., Li, C., Su, H., Zhu, J.: Prolificdreamer: High-fidelity and diverse text-to-3d generation with variational score distillation. arXiv preprint arXiv:2305.16213  (2023)

\bibitem{xu2023dream3d}
Xu, J., Wang, X., Cheng, W., Cao, Y.P., Shan, Y., Qie, X., Gao, S.: Dream3d: Zero-shot text-to-3d synthesis using 3d shape prior and text-to-image diffusion models. In: Proceedings of the IEEE/CVF Conference on Computer Vision and Pattern Recognition. pp. 20908--20918 (2023)

\bibitem{xu2023imagereward}
Xu, J., Liu, X., Wu, Y., Tong, Y., Li, Q., Ding, M., Tang, J., Dong, Y.: Imagereward: Learning and evaluating human preferences for text-to-image generation. arXiv preprint arXiv:2304.05977  (2023)

\bibitem{yi2023gaussiandreamer}
Yi, T., Fang, J., Wu, G., Xie, L., Zhang, X., Liu, W., Tian, Q., Wang, X.: Gaussiandreamer: Fast generation from text to 3d gaussian splatting with point cloud priors. arXiv preprint arXiv:2310.08529  (2023)

\bibitem{zhang2020nerf++}
Zhang, K., Riegler, G., Snavely, N., Koltun, V.: Nerf++: Analyzing and improving neural radiance fields. arXiv preprint arXiv:2010.07492  (2020)

\bibitem{zhang2019bertscore}
Zhang, T., Kishore, V., Wu, F., Weinberger, K.Q., Artzi, Y.: Bertscore: Evaluating text generation with bert. arXiv preprint arXiv:1904.09675  (2019)

\bibitem{efficientdreamer}
Zhao, M., Zhao, C., Liang, X., Li, L., Zhao, Z., Hu, Z., Fan, C., Yu, X.: Efficientdreamer: High-fidelity and robust 3d creation via orthogonal-view diffusion prior. arXiv preprint arXiv:2308.13223  (2023)

\end{thebibliography}
